\documentclass{WileyMSP-template}
\usepackage{todonotes}
\usepackage{lipsum}
\usepackage{xr}
\usepackage{parskip}
\usepackage{ragged2e}
\usepackage{amsmath}

\begin{document}

\pagestyle{fancy}
\rhead{\includegraphics[width=2.5cm]{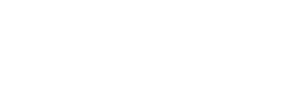}}

\title{MELEGROS: Monolithic Elephant-inspired Gripper with\\ Optical Sensors}

\maketitle

\author{Petr Trunin$^\dagger$}
\author{Diana Cafiso$^\dagger$}
\author{Anderson Brazil Nardin$^\dagger$}
\author{Trevor Exley*}
\author{Lucia Beccai*}

\begin{affiliations}
P. Trunin, D. Cafiso, A. B. Nardin, T. Exley, L. Beccai\\
Soft BioRobotics Perception\\
Istituto Italiano di Tecnologia (IIT)\\
Genova 16163, Italy\\
P. Trunin\\
The Open University\\
Affiliated Research Centre at Istituto Italiano di Tecnologia (ARC@IIT)\\
Istituto Italiano di Tecnologia\\
Genova 16163, Italy \\
E-mail: trevor.exley@iit.it, lucia.beccai@iit.it
\end{affiliations}


\begin{abstract}
\justifying
The elephant trunk exemplifies a natural gripper where structure, actuation, and sensing are seamlessly integrated. Inspired by the distal morphology of the African elephant trunk, we present MELEGROS, a Monolithic ELEphant-inspired GRipper with Optical Sensors, emphasizing sensing as an intrinsic, co-fabricated capability. Unlike multi-material or tendon-based approaches, MELEGROS directly integrates six optical waveguide sensors and five pneumatic chambers into a pneumatically actuated lattice structure (12.5 mm cell size) using a single soft resin and one continuous 3D print. This eliminates mechanical mismatches between sensors, actuators, and body, reducing model uncertainty and enabling simulation-guided sensor design and placement. Only four iterations were required to achieve the final prototype, which features a continuous structure capable of elongation, compression, and bending while decoupling tactile and proprioceptive signals. MELEGROS (132 g) lifts more than twice its weight, performs bioinspired actions such as pinching, scooping, and reaching, and delicately grasps fragile items like grapes. The integrated optical sensors provide distinct responses to touch, bending, and chamber deformation, enabling multifunctional perception. MELEGROS demonstrates a new paradigm for soft robotics where fully embedded sensing and continuous structures inherently support versatile, bioinspired manipulation. 

\end{abstract}

\section{Introduction}
\justifying
The elephant trunk is a remarkably versatile biological manipulator that integrates sensing and actuation within a jointless structure to grasp objects of many shapes, weights, and sizes. In the African elephant, the trunk tip has two asymmetric finger-like projections enable pinching, scooping, and supporting actions \cite{shoshani1997s,dagenaisElephantsEvolvedStrategies2021a,wuElephantTrunksForm2018a}. The absence of any division between the continuum arm and the tip, combined with distributed sensory feedback, supports smooth reaching, grasping, and highly dexterous manipulation tasks (\textit{e.g.}, ripping leaves from a wrapped branch). Unlike engineered systems that separate sensing, actuation, and structure, the elephant trunk links them through the integration of muscles, connective tissue, skin, and embedded mechanoreceptors. This arrangement allows the elephant to control movement and respond efficiently to contact with its environment, even with occluded vision during prehensile activity \cite{lopretiSensorizedObjectsUsed2023}. \textit{In this sense, the trunk is monolithic: structure, actuation, and sensing are inseparable.}

\begin{figure}[hbt!]
    \centering
    \includegraphics[width=1\linewidth]{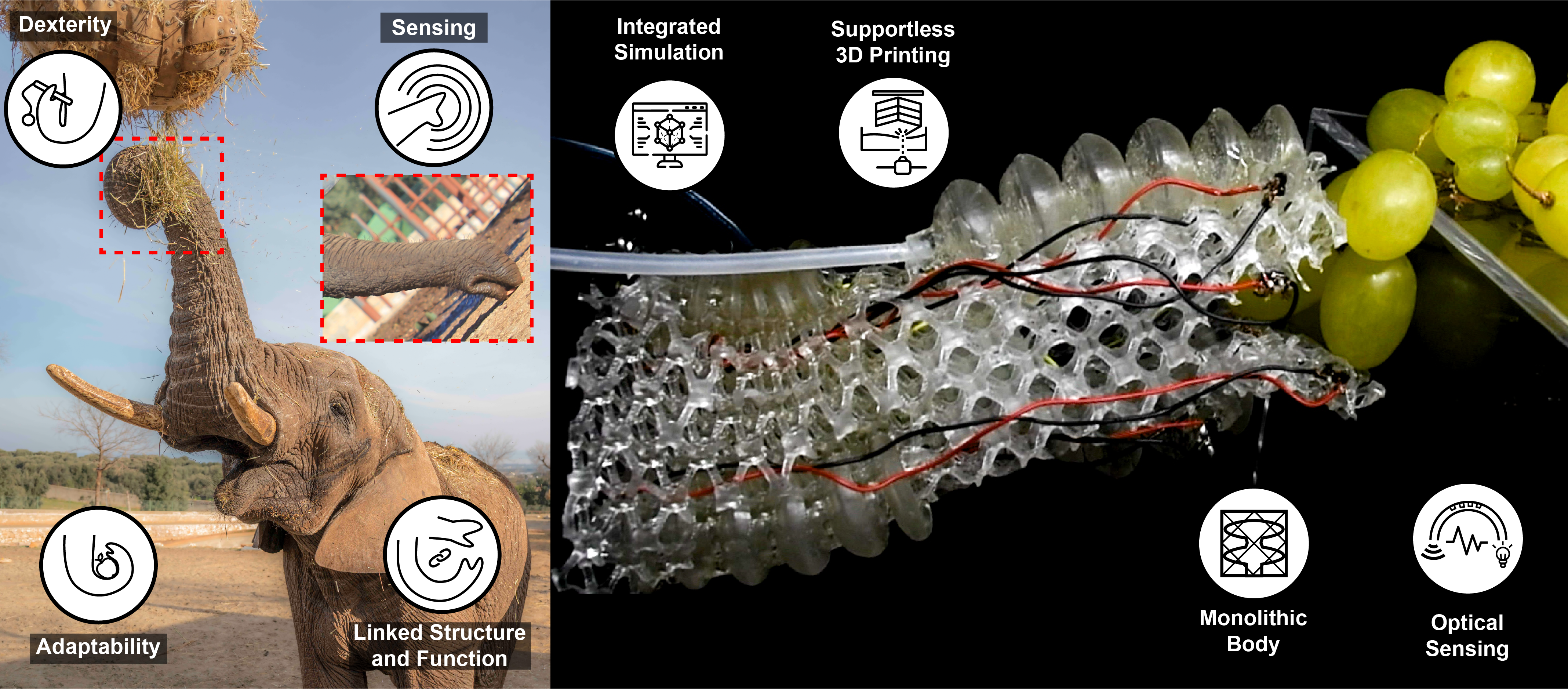}
    \caption{The MELEGROS concept inspired from the elephant trunk.  Left, photograph (courtesy of Matteo Montenero) of a male African elephant (ZooSafari, Italy), where the distal region of the trunk is highlighted. Right, the continuous structure of the artificial elongating gripper monolithically embedding optical sensing, fluidic actuation and lattice-based body. }
    \label{fig:comparison}
\end{figure}

This type of integration is still uncommon in robotic systems. Although today robots can deform and adapt to their surroundings through compliant materials and structures \cite{richUntetheredSoftRobotics2018,wangPerceptiveSoftRobots2018,chenSurveySoftRobot2025}, many are built by combining separate sensing, actuation, and structural elements \cite{shihElectronicSkinsMachine2020a,wangSensingExpectationEnables2024,xuOpticalLaceSynthetic2019}. This gap is related to a broader challenge in soft robotics: the lack of truly monolithic systems that combine all functional elements into a single material body. In particular, this problem stems from the fact that common transduction mechanisms, \textit{e.g.}, resistive \cite{liuTouchlessInteractiveTeaching2022,cafisoDLPPrintablePorousCryogels2024,soShapeEstimationSoft2021,jiDesignCalibration3D2023} and capacitive \cite{kimInherentlyIntegratedMicrofiberbased2024,huStretchableEskinTransformer2023,hashizumeCapacitiveSensingGripper2019, niuHighlyMorphologyControllableHighly2020} sensors, depend on conductive materials, with inherent mechanical characteristics (\textit{e.g.}, stiffness) different from the ones typically used for the robot's body. Moreover, sensing elements are often added post-fabrication. This leads to mechanical mismatches that can compromise compliance and induce failure under cyclic loading.  Recent efforts have achieved various degrees of actuator–sensor integration \cite{hanRecentAdvancesSensor2023}. For example, Truby et al. combined EMB3D printing with molding to produce soft somatosensitive actuators by injecting conductive ionogel into elastomeric matrices \cite{trubySoftSomatosensitiveActuators2018a}, while Xiao et al. fabricated a fully 3D-printed robotic hand incorporating soft capacitive sensors via dual-extrusion of dielectric and conductive silicones \cite{xiaoFully3DPrintedSoft2025}. In addition to transduction strategies, the pressure feedback from fluidic channels has been leveraged to maintain material uniformity \cite{trubyFluidicInnervationSensorizes2022}, though at the expense of increased design complexity (thin hollow channels) and potential performance trade-offs between sensing and actuation. Although these approaches represent meaningful advances toward the monolithic approach, they still rely on multiple materials, involve elaborate fabrication workflows, and restrict design versatility. To fully eliminate material mismatches and post-assembly procedures, a solution is to build sensors with the same material as the actuators and the robot body, a goal attainable by implementing transducers that exploit the optical, rather than the electrical, conductivity of sensing materials. In fact, the monolithic integration of optical sensing set just one requirement, \textit{i.e.}, the material of both sensors and robot must be transparent to light. This feature, even if challenging, is still less constraining than those required from other transduction mechanisms, which may imply the addition of functional fillers (\textit{e.g.}, magnetic, electrically-conductive) stiffening the robot and creating bi-material interfaces. As a first step, our previous work introduced the Monolithic Perceptive Unit (MPU): a fully 3D-printed lattice cell in which the constituent elastomer itself functions as an optical sensor \cite{truninDesign3DPrinting2025b}.

Lattice architectures have emerged as a promising alternative to bulk soft bodies. Although bulk material functionalization is possible, it often comes at the expense of mechanical performance \cite{imanianDesign3DPrinting2025}. In contrast, lattice architectures retain the softness of the bulk material while offering internal pathways and anchor points, which simplify the integration of actuators and sensors and enable support-free fabrication by creating \textit{in situ} supports during 3D printing. For example, tendon-driven lattices have been used to reproduce musculoskeletal behaviors by routing cables through the structure \cite{guanLatticeStructureMusculoskeletal2025,schouten3DPrintableGradient2025b}. Alternatively, we have previously introduced lattice-embedded actuators which can achieve bending and jointless behavior \cite{joeJointlessBioinspiredSoft2023a}. In this work, an IWP-TPMS (triply periodic minimal surface) lattice is adopted. While the underlying topology defines the deformation modes, the stiffness -in addition to the intrinsic material properties- depends on the cell dimensions. Since the lattice serves as a compliant medium for the embedded actuators, and it must enable the gripper to extend, bend, and conform around objects during grasping, a low bending stiffness of the chosen lattice configuration is pursued.

The draw of the proposed monolithic method in soft robotics lies in simplifying the design process: using a single-material body that combines actuation and sensing without post-processing. However, few materials are capable of delivering all required functionalities (\textit{i.e.}, flexibility, printability, sensing) while remaining compatible with streamlined fabrication. Recent advances in commercial 3D printing \cite{yaoJAMMitMonolithic3DPrinting2025} have enabled one-step processes to create hybrid systems, yet fully printed monolithic soft systems are still rare.  

Inspired by the morphology and behavior of the distal region of the African elephant trunk \cite{dagenaisElephantsEvolvedStrategies2021a, lopretiSensorizedObjectsUsed2023}, and enabled by an architected design, we introduce the MELEGROS concept: a \textbf{M}onolithic \textbf{Ele}phant-Inspired \textbf{Gr}ipper with \textbf{O}ptical \textbf{S}ensors (Figure \ref{fig:comparison}). The system is built from a soft lattice with smoothly connected bladder-shaped actuators, which not only allow the system to elongate, compress, and bend, but act as the body of both gripper and soft optical sensors. The design does not aim to mimic natural muscular arrangements, but rather focuses on functional integration of actuation and sensing. To achieve this objective, and building on our recent results in simulating soft lattice structures \cite{nardin_exploring_2025}, our method is based on a workflow linking design and fabrication through simulation in SOFA (Simulation Open Framework Architecture)~\cite{duriez_realistic_2006, payan_sofa_2012} to design and position the soft optical sensors. The output is a fully-integrated design that is fabricated via a single 3D-printing process. In this work, we focus on investigating the sensing functionality by targeting the discrimination of exteroceptive from proprioceptive information during grasping tasks. We show how the specific monolithic architecture enables MELEGROS to perform enveloping grasps and delicate pinching maneuvers (extending its functionality beyond simple parallel-jaw closure) and to reach and bend independently in an intrinsic workspace, where a specific object can be reached from multiple directions before being grasped, similar to the natural model.

\section{Results}
\subsection{Design of the Gripper}

The design process begins with two key steps that establish the basis for monolithic integration. First, the lattice geometry is selected and characterized through mechanical testing (Figure \ref{supp-fig:lattice}), providing homogenized stiffness values for use in simulation. Second, a half-embedded actuator is fabricated to validate printability and assess kinematics during actuation (Figure \ref{supp-fig:halfembedactuator}). These preliminary results informed the integration of actuators within the lattice and guided the simulation of the complete gripper presented in the following sections. 

To address monolithic integration, we implement an iterative workflow that links design (of structure, sensing, actuation) and fabrication through simulation (Figure \ref{supp-fig:workflow}). At the core of the monolithic approach is the lattice which has three main roles. First, it enables monolithic fabrication of sensors and actuators. Its struts and voids provide a scaffold for the waveguides and the actuators, which can be printed all together without supports. This leads to a monolithic architecture and reduces post-processing. Second, it enables full compliance and adaptability during grasping. Owing to its TPMS geometry, the lattice can deform laterally, allowing the gripper to conform around objects of varying shapes (e.g., cube, sphere, star) and sizes (in the range of 12.5 to 25 mm) while maintaining contact along multiple surfaces. This flexibility supports both enveloping grasps and delicate pinching, enabling interaction with a wider variety of objects than simple parallel-jaw closure would allow. Third, it transmits the deformation of the embedded longitudinal actuators, without critically limiting the elongation of the whole gripper and enabling reaching.

An IWP-type TPMS geometry (Equation \ref{supp-eq:TPMS}) is adopted for the lattice, which reduces bending stiffness (w.r.t. a same-size bulk solid) while maintaining structural continuity, allowing the distal arm and the tip to deform as a single continuous unit, mimicking the bioinspired integration observed in elephant trunks. The lattice is designed to be compliant and to enable supportless 3D printing, while accommodating embedded optical sensing and serving as the body of the gripper. Strut thickness is chosen (1.5 mm) to maintain flexibility while supporting the integrated waveguides. The selected cell size is determined through iterative prototyping to achieve printability with sufficient compliance (Young's modulus $<$ 1 MPa) \cite{majidiSoftMatterEngineeringSoft2019, polygerinosSoftRoboticsReview2017}. The chosen cell size (12.5 mm) determines the minimum object size that can be grasped without slipping within the lattice air cell ($\sim$8 mm void) while attached to the actuators (Figure \ref{supp-fig:dimensions}). To evaluate its mechanical response, cubic lattice samples are subjected to compression stress-strain tests, demonstrating nonlinear stiffness and recoverable deformation (Figure \ref{supp-fig:lattice}). These results are used to inform the simulations  (described in later sections), where the response is linearized up to 40\% strain and approximated by an effective Young’s modulus of $\sim$12 kPa.  This modulus is applied to the design's bulk volume envelope (Figure \ref{supp-fig:dissection}) as the homogenized material property for the lattice.

Actuators capable of elongation and contraction are selected, so when they are printed half-embedded by the lattice, the anchor points introduce strain-limiting behavior, which enables the actuators to be repurposed for bending (Figure \ref{supp-fig:halfembedactuator}). Bladder-like chambers are chosen because their circular cross-section allows direct attachment to the TPMS nodes, while their convoluted geometry accommodates local expansion at lattice anchor points and compression of the intervening voids. The pneumatic actuators are implemented as bladder-like chambers arranged in series and co-designed with the lattice anchor points to ensure symmetric deformation between chambers. 

\begin{figure}[t!]
    \centering
    \includegraphics[width=0.85\linewidth]{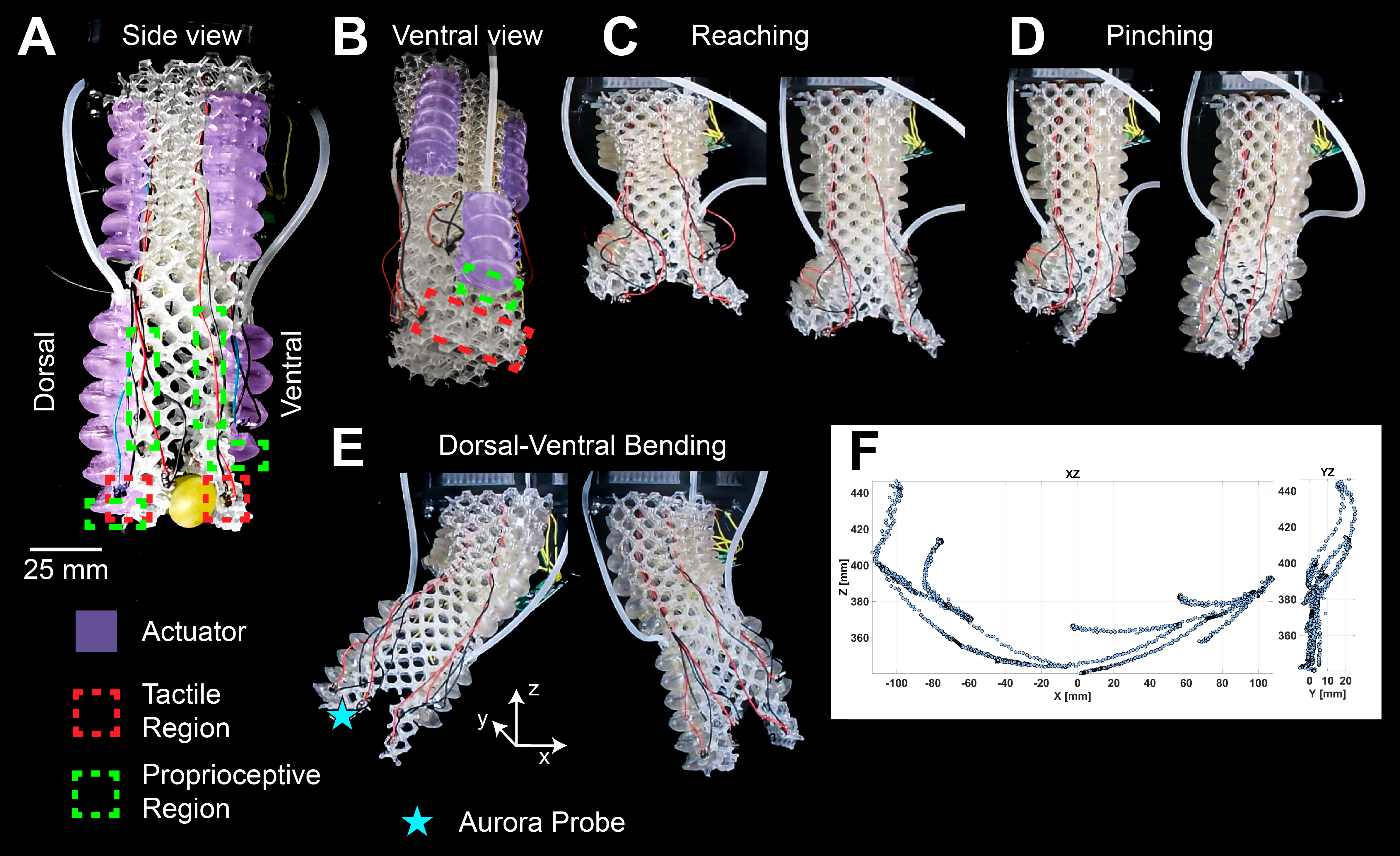}
    \caption{Overview of MELEGROS design and motions. (A) Side and (B) ventral views of MELEGROS depicting bladder-like actuators in purple (dorsal and ventral fingers have an actuator with 6 and 4 chambers, respectively; proximal region integrates three 6-chamber actuators), with sensor regions of interest indicated by red and green rectangles. MELEGROS during (C) reaching, (D) pinching, and (E) dorsal-ventral bending. A star indicates where the electromagnetic probe (AURORA, Northern Digital Inc., Canada) was placed for defining the (F) workspace boundaries defined as $x \in [-113,107]\ \text{mm}, \; y \in [-5,25]\ \text{mm}, \; z \in [340,446]\ \text{mm},$ captured sweeping across dorsal and ventral bending.}
    \label{fig:design}
\end{figure}


The gripper design is conceptually inspired by the elephant trunk tip, whose asymmetric finger-like projections achieve versatile grasping behaviors (Figure \ref{supp-fig:melegrosdimensions}). Similarly, our system integrates two opposing lattice-supported actuators at the distal end of the structure, with the dorsal and ventral fingers having 6 and 4 chambers, respectively (Figure \ref{fig:design}). Each actuator is embedded along the lattice struts in such a way that pressurization induces bending toward the central axis of the gripper. Three 6-chamber actuators are placed proximal (\textit{i.e.}, at the base) of our design to achieve contraction, elongation, and bending. These proximal actuators are arranged with one on the dorsal side along the central axis, and two on the ventral side resembling the natural ‘ventral ridge’ \cite{schulzElephantsDevelopWrinkles2024a,deiringerFunctionalAnatomyElephant2023a}. The objective herein is not to mimic the natural muscular arrangements \cite{dagenaisElephantsEvolvedStrategies2021a}, rather the placement of the actuators is determined to achieve some bioinspired behavior like the reaching, pinching, and dorsal-ventral bending, enabling independent reach and bending motion modalities. This way MELEGROS can move in a space suitable for approaching objects from multiple directions without their precise pre-alignment with the tip, as will be shown later.

Functionally, the asymmetric actuator arrangement is intended to mimic some basic motions of the distal trunk (Figure \ref{fig:design}). When the proximal actuators are driven together, the surrounding lattice compresses and expands to achieve reaching (Figure \ref{fig:design}C). These proximal actuators are also capable of moving independently, resulting in dorsal and ventral bending (Figure \ref{fig:design}E). The finger actuators can be driven asymmetrically, one finger bending inward while the other remains passive, enabling pinch-like grasping (Figure \ref{fig:design}D) of multiple objects (Figure \ref{supp-fig:objects}). Additionally, MELEGROS (m = 132 g) can drive its fingers together to grasp and lift an object weighing twice its own mass (cylinder: L = 120 mm, d = 30 mm, m = 264 g) (Video S1).


\subsection{Sensing design and simulation}
\begin{figure}[b!]
    \centering
\includegraphics[width=0.85\linewidth]{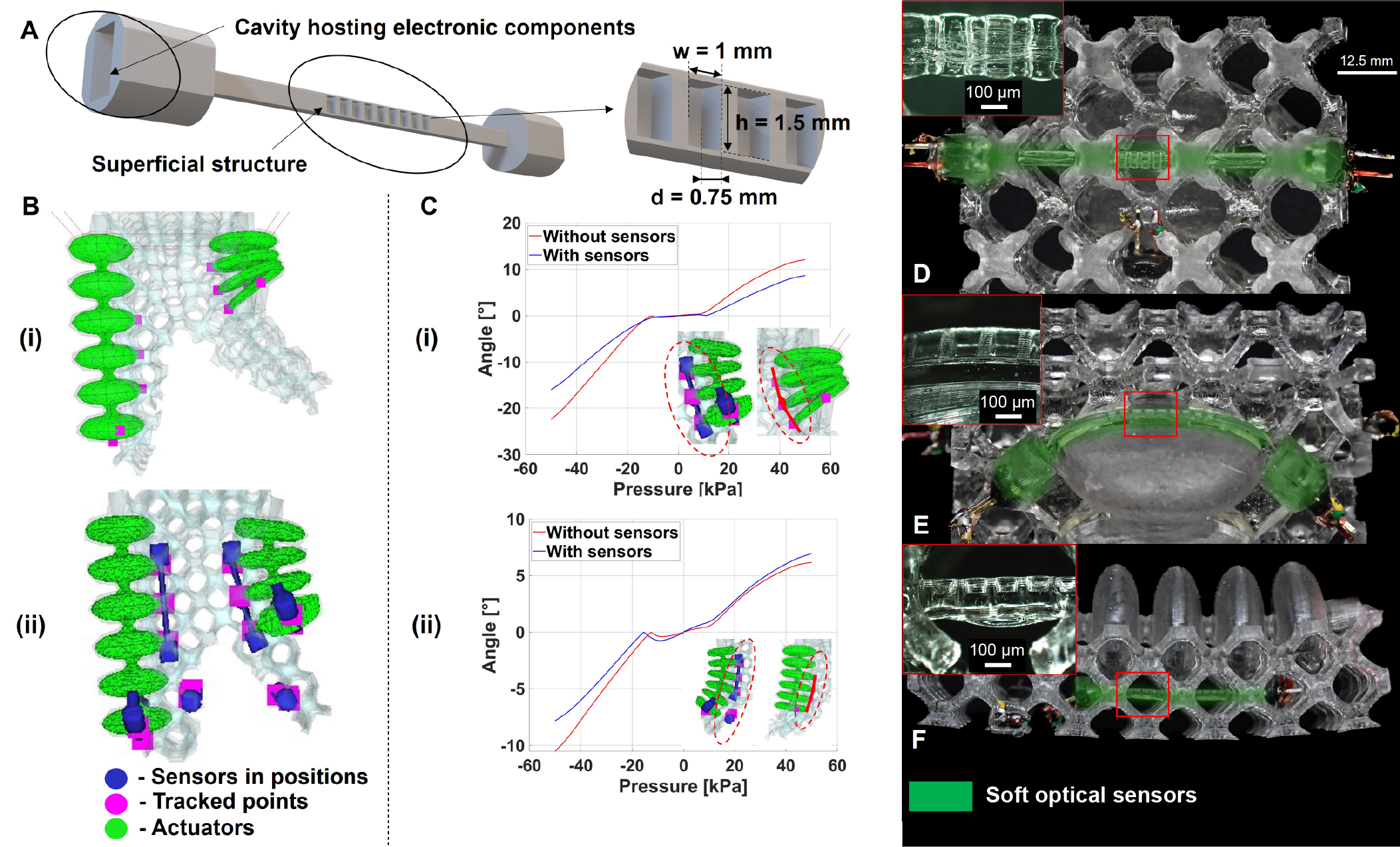}
    \caption{Design, simulation, and placement of optical sensors. (A) Design of the optical sensors and their superficial structures (width = 1 mm, height = 1.5 mm, depth = 0.75 mm). (B) Simulations with tracked points within (i) Model A: sensor regions of interest, and (ii) Model B: incorporated sensor structures. (C) Simulated finger bi-directional bending angles from -50 kPa to 50 kPa to inform design and placement for (i) ventral bending and (ii) dorsal bending sensors. (D) Tactile sensor (37.5 mm long), (E) pressurized actuator sensor (25 mm long), and (F) finger bending sensor (37.5 mm long) highlighted in digital microscope images.}
    \label{fig:sensors}
\end{figure}

Sensing in MELEGROS is focused on measuring bending of the fingers, deformation in fluidic chambers, and tactile response. Optical waveguide sensors are used (which we introduced in a previous work \cite{truninDesign3DPrinting2025b}): 3D-printed polymer waveguides incorporating a surface pattern (\textit{i.e.}, a series of wells) that amplify bending sensitivity (Figure \ref{fig:sensors}A). In brief, an LED emits light into the waveguide towards a photodetector. Bending increases optical losses, primarily via scattering from the patterned surface, thereby reducing the received intensity and producing a corresponding decrease in the photodetector voltage. The general waveguide design for both proprioceptive and tactile sensors is provided in Figure \ref{supp-fig:sensorbehavior}, aimed at producing a sensing response due to actuator motion and external contact, respectively.

In comparison to previous work, a new parametrization of the sensor design is implemented to enable monolithic fabrication. Specifically, adjustments are made to the thickness of the waveguide and the geometry of the surface pattern (Figure \ref{supp-fig:sensingdesign}A). Sensor thickness is constrained by two factors: waveguide printability and the printer’s nominal resolution (50 $\mu$m). Additionally, the waveguide must be sufficiently thick to accommodate wells of adequate depth without compromising durability under repeated loading. Given that the minimum well depth to obtain a reliable print is 0.5 mm, the waveguide thickness is set to 1.5 mm, which also establishes the aforementioned lattice strut thickness. The dimensions of the surface pattern are defined by maximizing linearity and sensitivity, by using COMSOL Multiphysics simulations (Figure \ref{supp-fig:sensingdesign}A). Parametric sweeps of the pattern geometry yield a well depth d = 0.65 mm and width w = 1.0 mm. Further details are provided in Figure \ref{supp-fig:sensingdesign}B. 
Then, sensor placements are established to ensure output signal fidelity while preserving mechanical compliance. This is guided by simulated position trajectories, targeting regions that consistently undergo contact with objects during typical finger motions (for tactile sensing) and regions that experience bending trajectories independent of contact (for proprioception). 

The methodology begins with defining an initial gripper design (informed by the lattice printability trials) in simulation. Observations of the half-embedded actuator movements provide early feedback on the potential design and placement of the optical waveguides in the simulated gripper. Then, the sensing regions of interest are defined (Model A: Figure \ref{supp-fig:sensor_position_model}). The following steps consist of defining some points in these regions based on waveguide geometry (Figure \ref{fig:sensors}B(i)) and measuring their position trajectory during actuation (from -50 kPa to 50 kPa). If the displacement is sufficient to warrant a sensing response, then waveguides are added to the robot model within the simulation environment (Model B: Figure \ref{supp-fig:sensor_integrated_model} and Figure \ref{fig:sensors}B(ii)). Consequently, the simulation framework Model B is used until some specific conditions are fulfilled. First of all, the variation of angle in the actuator-waveguide system must confirm that it is possible to get a sensing response in the real prototype. Additional consideration for the sensors’ placements must be given to the printability (using the lattice nodes as support elements of the waveguides). Indeed, if the waveguide behavior is decoupled (\textit{e.g.}, tactile waveguide bending $<$ 5$^\circ$ during finger bending shows a small influence of general movements on tactile sensors) and if the integrated waveguides do not significantly change the mechanical behavior of the gripper (\textit{e.g.}, change in bending angle $<$ 5$^\circ$ doesn’t affect the task performance), the iterative process is finished. This analysis, critical for distinguishing sensing signals deriving from touch from those induced by gripper motion, is made possible by the monolithic architecture, which removes material interfaces between sensors and actuators and thereby reduces model uncertainty.


According to these requirements, tactile and proprioceptive regions of interest are defined as depicted in Figure \ref{fig:design}A. and the finalized sensor layout is shown in Figure \ref{fig:sensors}D-F. Specifically, a total of six soft optical waveguides are placed as follows: (i) a tactile sensor is on the inner surface of each finger (Figure \ref{fig:sensors}D); (ii) a sensor is on the last chamber of each finger actuator, to sense the fluidic deformation (Figure \ref{fig:sensors}E); and, (iii) a sensor is along the length of each finger, to sense the bending motion (Figure \ref{fig:sensors}F). Because the dorsal and ventral fingers contain actuators with six and four chambers, respectively, the sensor lengths follow this asymmetry, with the ventral finger incorporating a shorter bending sensor.

Herein, we explain how this layout allows for satisfying all the aforementioned requirements. The bending angles of the sensor trajectories observed in the simulation (Figure \ref{fig:sensors}C) reach up to 40$^\circ$, which we previously demonstrated to be sufficient for generating a sensing response for this type of sensor \cite{truninDesign3DPrinting2025b}. This is further validated in Section 2.3, where the sensor response is characterized in the half-embedded actuator integrating the three sensor types presented in this work. To minimize cross-talk, bending sensors are placed along segments of observed curvature (close to the actuators, as shown in Figure \ref{fig:design}) that remain contact-free from the objects due to their position inside the lattice. Tactile sensors are instead positioned on the inner finger surfaces where contact is expected during grasping.

The results reported in Figure \ref{supp-fig:angle_sensors36} confirm that the simulated bending of tactile waveguides is negligible during the opening and closing phases of the gripper, providing strong hints on the possibility to decouple proprioceptive and tactile feedback in the real prototype. Finally, the interaction between the compliant lattice and the regions stiffened by sensor inclusion was represented to verify the effect of the waveguides on the mechanics of the gripper. The presence of sensors (waveguides) locally increases the rigidity of the structure and thereby alters its deflection response (Figure \ref{fig:sensors}B). Figure \ref{fig:sensors}C illustrates this effect: when sensors are not included, the actuator exhibits larger angular deflections under the same pressure, whereas adding the sensor domains leads to slightly reduced angles. This outcome confirms the expected result that sensors contribute mechanical stiffness to the system, an effect that must be taken into account to correctly predict their position trajectories.

To further examine the sensing regions, 2D projections of sensor-point trajectories are analyzed (as described in the Supplementary Information). For example, for Model A, the ventral finger modes \textbf{open2} and \textbf{close2} (Figure \ref{supp-fig:spm_open2} and \ref{supp-fig:spm_close2}) reveal that, even in the absence of embedded sensors, the monitored points trace smooth and consistent paths over the full actuation cycle, confirming their suitability as proprioceptive sites. For Model~B, the corresponding trajectories during the simultaneous-finger modes \textbf{open} and \textbf{close} (Figure \ref{supp-fig:sim_open} and \ref{supp-fig:sim_close}) further demonstrate that integrating the sensors has negligible impact on overall motion while preserving distinct trajectories for angular readout. 

Eventually, the final monolithic prototype is successfully printed. The integration between waveguides, lattice, and actuators is shown in Figure \ref{fig:sensors}D-F. The optical waveguides (highlighted in green) are co-printed with the gripper, using the lattice and the actuator chambers as \textit{in situ} supports. Magnified insets display the surface patterning across sensor types, confirming high fidelity in small-feature fabrication.

\subsection{Characterization and Sensing Response}

\begin{figure}[b!]
    \centering
    \includegraphics[width=0.85\linewidth]{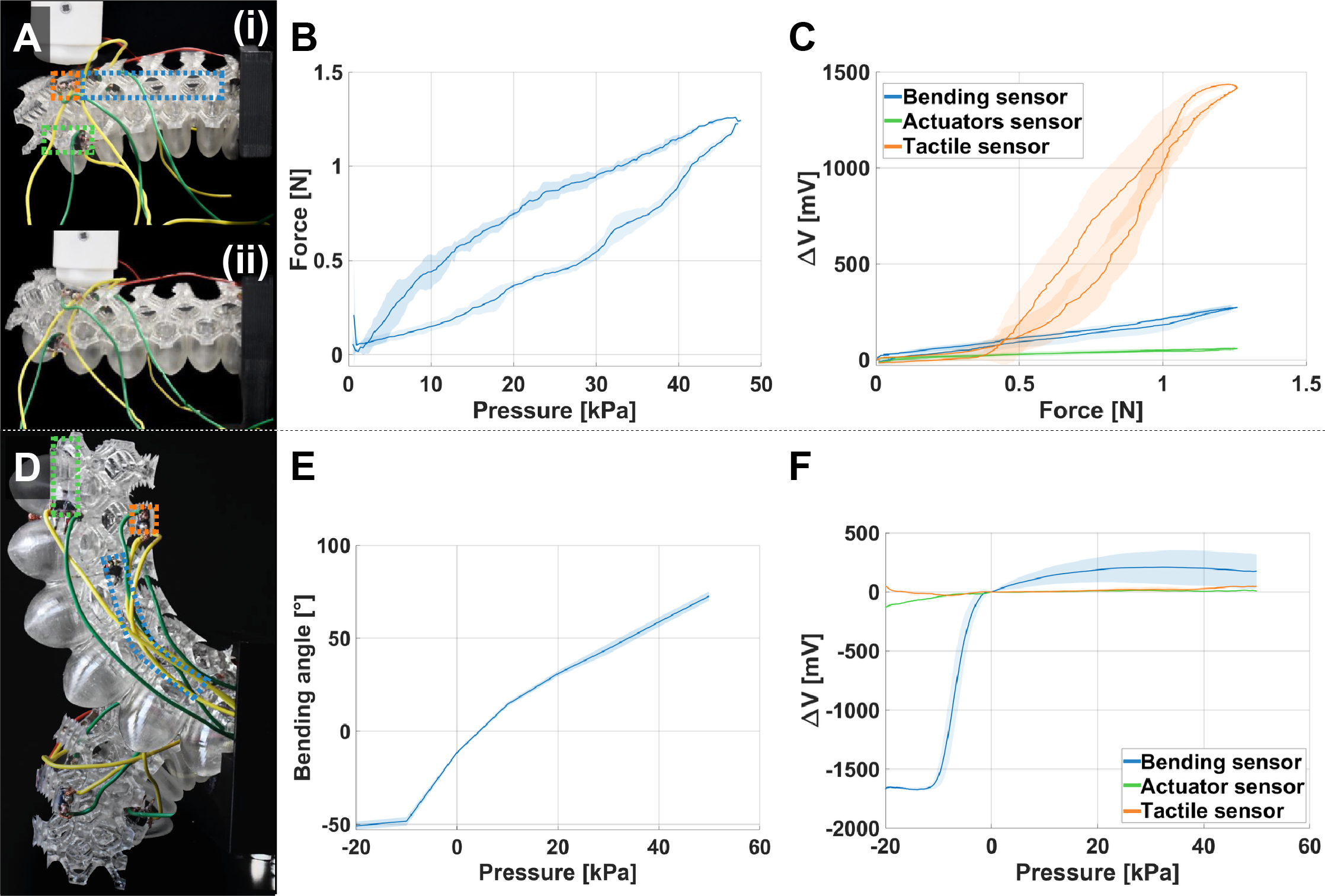}
    \caption{Characterization of sensorized half-embedded actuators.  (A) Blocking force tests of the half-embedded bladder-like actuators at (i) rest, and (ii) actuated state. (B) Resultant force data from the triaxial load cell, and (C) corresponding sensor responses.  (D) Free bi-directional bending of the half-embedded actuator with (E) recorded angles were captured from -20 kPa to 50 kPa in ImageJ, and (F) corresponding sensor responses. Sensor positions are outlined in panels A and D in their respective colors and detailed images of the final version of the sensors are shown in Figure \ref{fig:sensors}D-F. Shaded areas of graphs represent $\pm$1 standard deviation (N = 3).}
    \label{fig:characterization}
\end{figure}

Following the simulation informing the final optical sensor placement, two prints can be fabricated. The first is the sensorized version of the half-embedded actuator, used for the actuator's characterization and preliminary check of sensors' response, since its movement resembles that of MELEGROS. The second is MELEGROS, which is used for grasping applications to assess sensor fidelity. 
For the sensorized half-embedded actuators, experimental tests are conducted within a pressure range of –20 to 50 kPa, as lower pressures (below –20 kPa) did not produce additional bending. Within this range, bending angles and blocking forces are observed to scale with applied pressure (Figure \ref{fig:characterization}). Half-embedded actuators achieve free bending angles from –50$^\circ$ (–20 kPa) to 75$^\circ$ (50 kPa). Under positive pressure, half-embedded actuators expand to produce a blocking force of 1.25 N (50 kPa).

\begin{figure}[b!]
    \centering
    \includegraphics[width=0.7\linewidth]{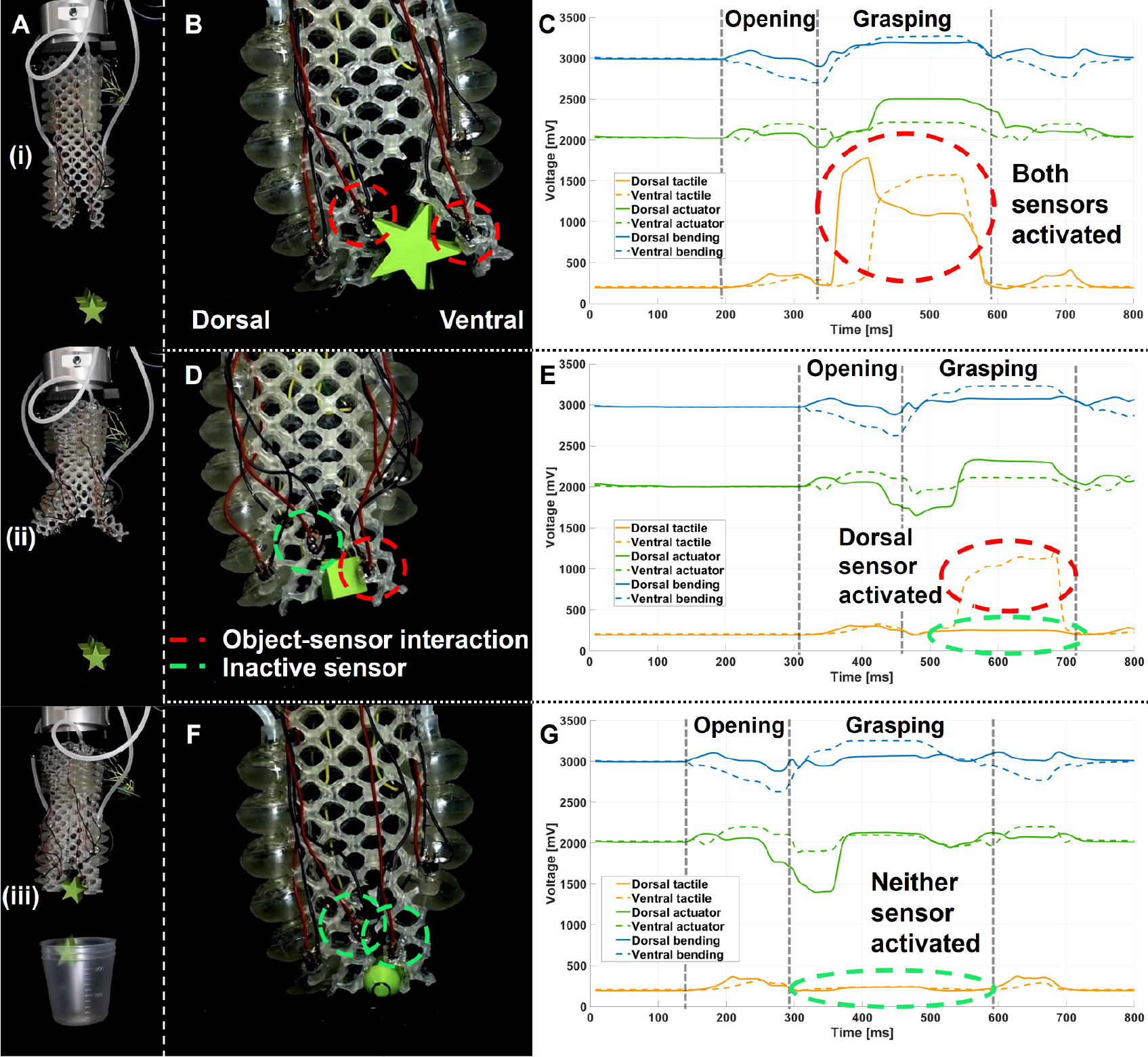}
    \caption{Object Grasping and Sensor Response in MELEGROS. (A) MELEGROS (i) approaching, (ii) grasping, and (iii) releasing multiple objects (three stars, 12.5 mm). Grasping scenarios where sensors are activated in (B-C) both fingers (two stars, 25 mm), (D-E) only the ventral finger (three cubes, 12.5 mm), and (F-G) neither finger (three spheres, 12.5 mm).}
    \label{fig:graspingobjects}
\end{figure}

Sensor responses are collected on the same half-embedded actuators under these blocking-force and free-bending tests (Figure \ref{fig:characterization}). Under blocking-force conditions (Figure \ref{fig:characterization}A-C), the tactile sensor (orange) produced the largest response ($\Delta$V $\approx$ 1.4 V), the bending sensor (blue) an intermediate response ($\Delta$V $\approx$ 0.25 V), and the actuator sensor (green) the smallest ($\Delta$V $\approx$ 0.06 V); each dataset comprises three cycles. During free bending experiments (Figure \ref{fig:characterization}D-F), the bending sensor exhibited the largest response ($\Delta$V $\approx$ 2.0 V), and its voltage trajectory resembles the measured bending angle. The distinct response profiles indicate effective decoupling between tactile and proprioceptive information. 

Following the characterization of the previous lattice/actuator/sensors structure, MELEGROS is evaluated through grasping trials to assess sensor responses during these tasks. The asymmetric structure of the tip enables a bioinspired grasping technique, in which the ventral finger can remain passive, while the dorsal scoops objects. Due to this capability, MELEGROS can successfully grasp and lift not only the aforementioned heavy (double its weight, 264 g) and elongated objects (cylinder: L = 120 mm, d = 30 mm), but also small objects of various shapes, such as spheres, cubes, and stars with different dimensions (d = 12.5, 25 mm) (Figure \ref{supp-fig:objects}). Moreover, grasping of multiple objects is shown and achieved owing to the adaptability of the lattice structure.

Furthermore, MELEGROS is installed on a robotic arm (UR5e, Universal Robots, Denmark) to perform pick-and-place tasks with different objects (Video S2) while recording the response of the sensors. The sensing response is influenced by the different movements and sizes of the objects (Figure \ref{fig:graspingobjects}). The behavior of the sensorized gripper can be summarized as follows: from a relaxed state, it contracts along its axis and opens the fingers (Figure \ref{fig:graspingobjects}A(i, ii)), the robot arm approaches the target, MELEGROS elongates, and the fingers close. These phases elicit distinct signals: actuator and bending sensors report chamber pressurization and finger deformation during opening/closing, whereas tactile sensors respond primarily upon object contact. 
Following this, the robot arm moves above a cup, MELEGROS opens, and the object is released (Figure \ref{fig:graspingobjects}A(iii)). 
Across experiments, all the sensors exhibit reproducible patterns. Owing to gripper asymmetry, tactile responses vary with respect to object size (Figure \ref{fig:graspingobjects}B-G): larger items (e.g., 25 mm star shapes) activate both tactile sensors on the dorsal and ventral fingers (Figure \ref{fig:graspingobjects}B-C), whereas smaller items activate only one sensor or none (Figure \ref{fig:graspingobjects}D-G). During the grasping of three small cubes (Figure \ref{fig:graspingobjects}D-E), only the ventral finger comes in contact with the objects and is activated. For the case of the small spheres (Figure \ref{fig:graspingobjects}F-G), neither tactile sensor are activated, as their capture does not laterally deform the lattice.

\subsection{MELEGROS in action}

\begin{figure}[htb!]
    \centering
    \includegraphics[width=0.7\linewidth]{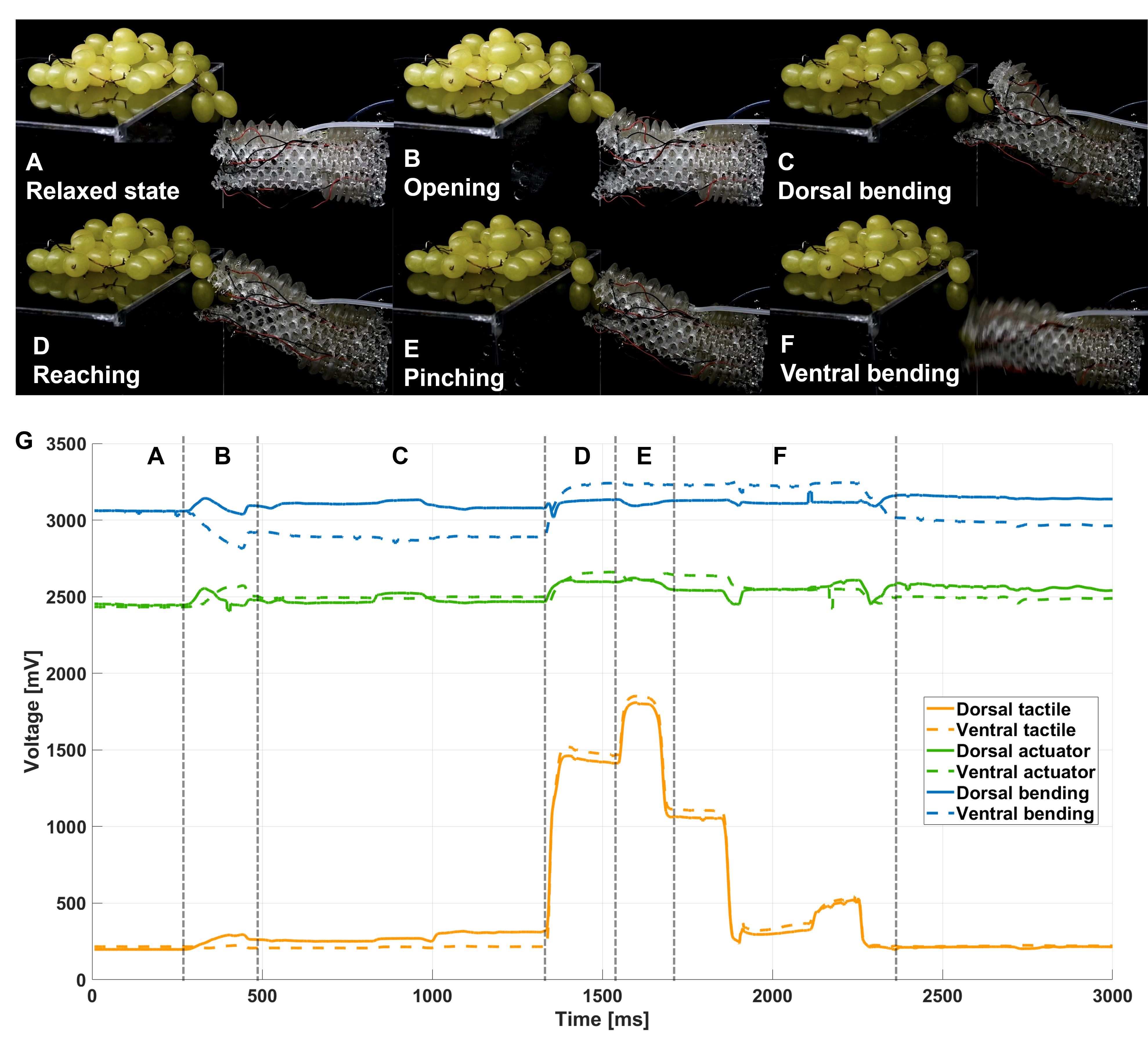}
    \caption{MELEGROS is approaching a bunch of grapes from the side, with corresponding sensor data for all six sensors. MELEGROS in (A) relaxed state, (B) opening, (C) dorsal bending, (D) reaching towards and coming into contact with the grape, (E) pinching the grape, and (F) ventral bending (Video S3).}
    \label{fig:grapes}
\end{figure}

Conventional soft grippers primarily execute pinch grasps and function solely as end-effectors. In contrast, MELEGROS can bend and extend independently, providing improved freedom of motion to approach an object from multiple directions and to grasp it without requiring a precise pre-positioning of the same. Crucially, the gripper performs these manoeuvres while delivering motion-specific sensory readouts.
Figure \ref{fig:grapes} and Video S3 depict dexterous reaching and grasping of delicate fruit (grapes). In this scenario, the gripper is positioned near the target and opens from a contracted state (Figure \ref{fig:grapes}A,B), performs dorsal bending (Figure \ref{fig:grapes}C), reaches by elongating ventral actuators to contact the grape (Figure \ref{fig:grapes}D), pinches the grape (Figure \ref{fig:grapes}E), and then bends ventrally to detach/pick the grape from the stem before repositioning and opening to release (Figure \ref{fig:grapes}F).

Moreover, the sensors respond to each phase of the motion, as shown in Figure \ref{fig:grapes}G. In the relaxed state (A), the sensory signals remain constant. During opening (B), bending and actuator sensors in each finger are activated. When the base bends dorsally without finger movement (C), sensor outputs remain unchanged. In the reaching and closing phase (D), both bending and actuator sensors are activated again, but with a distinct response compared to the "simple" closing (empty state), while the tactile sensors confirm contact with the object. During pinching and picking (E), bending and actuator sensors remain constant, but tactile sensors register a sharp peak as force is applied to detach the grape, followed by an immediate drop once it is removed. Finally, during ventral bending and opening (F), bending and actuator sensors provide steady feedback during base bending, then return to the values associated with opening. Tactile responses decrease after picking, reflecting the reduced contact force, and ultimately return to relaxed-state values.

\section{Discussion and Conclusion}
This work advances soft robotics design by demonstrating a monolithic strategy in which actuation and sensing are co-designed, co-fabricated, and co-located within a single lattice body. Unlike conventional approaches, and existing monolithic approaches \cite{zhaiDesktopFabricationMonolithic2023} that depend on trial-and-error prototyping, we establish a simulation-reinforced workflow in which tracked position trajectories predict how sensors and actuators interact within the body. Noteworthy, we use simulation as a design tool (rather than a post hoc validation step) to address a long-standing challenge in soft robotics: the decoupling of tactile and proprioceptive feedback within a single compliant body. 
The main aspects of our work deserve emphasis as generalizable contributions. 
The presented monolithic approach joins sensing and structure, and proves the possibility of building continuum manipulators where the arm and the end-effector form a continuous structure. Mimicking such basic feature of the natural distal trunk gives rise to a new concept of a soft robotic manipulator. 

Indeed, the proposed lattice architecture enables support-free printing of complex compliant features and internal cavities, eliminating sacrificial materials and post-processing. Then, integrating actuators within the compliant lattice body, MELEGROS is able to reach, bend, grasp, and release objects (singular or multiple) within a workspace.  The results of the preliminary grasping and manipulation experiments are promising and suggest that our work goes beyond other approaches resembling the elephant tip, which rely on rigid robotic arms to navigate the workspace \cite{chenBioinspiredStrategyInspired2024,washioDesignControlMultiModal2022a,wangReliableDamagelessRobotic2025a}. Moreover, crucially, tactile and proprioceptive information are provided by the body itself, serving as the transduction medium, removing internal interfaces and collapsing the distinction between “robot” and “sensor”, reinforcing our central premise.

Finally, but not less important, reliable simulation is enabled by the material continuity and gives important insight on sensing region of interest trajectories to tune sensors' geometry and placement before printing, turning what is usually iterative trial-and-error into a targeted, more directed design step. Therefore, only a few different prints in total are required in this entire process: the half-embedded actuator and lattice before sensor placement, and the actuator for characterization and gripper after sensor integration.

Building on these results, the next phase will advance MELEGROS toward robust closed-loop manipulation. The embedded sensors enable feedback control, a defining capability for robotic systems. Priorities will include quantitative investigation of payload limits, implementation of closed-loop control with phase-resolved feedback, and long-duration testing to characterize durability and drift. In parallel, graded materials \cite{schouten3DPrintableGradient2025b} and interpenetrating lattices \cite{whiteInterpenetratingLatticesEnhanced2021a, guanLatticeStructureMusculoskeletal2025} within the robot's body will be explored to extend functionality and deepen bioinspired design. In particular, these strategies open the possibility of introducing stiffness gradients within a monolithic body. Such graded architectures would support multiple movement and sensing modalities, moving beyond current manipulators while echoing the functional and morphological continuity of the natural trunk. As the field moves toward hybrid soft-rigid grippers for enhancing payload capacity \cite{chenSoftrigidCouplingGrippers2023,zhaoEnhancingGraspingDiversity2025}, it is noteworthy that MELEGROS achieves a payload of twice its system weight while remaining a fully monolithic soft structure. This highlights the need for further systematic evaluation of payload capabilities in soft monolithic systems across different application contexts. Finally, scaling via larger build volumes or modular printing will support a full trunk with further elephant-inspired modalities such as wrapping \cite{dagenaisElephantsEvolvedStrategies2021a} and twisting.

\section{Experimental Section}
\threesubsection{Material and Printing Parameters}\\
All soft structural components were fabricated using Elastic 50A Resin (Formlabs, United States), a flexible photopolymer suitable for high-strain applications. Samples were printed using a Form 4 (Formlabs, United States) stereolithography (SLA) printer with a layer thickness of 0.1 mm. Exposure parameters were as follows: perimeter, model, and support fill regions received an energy dose of 38.40 mJ/cm\textsuperscript{2}, whereas the light intensity was set as 11.5 mW/cm\textsuperscript{2}.

Post-processing was carried out using the standard Formlabs protocol. Printed parts were washed in isopropyl alcohol for 20 minutes using a Form Wash station, followed by UV curing at 70 $^\circ$C for 30 minutes in a Form Cure unit to ensure complete polymerization.




\threesubsection{Characterization Specimens and Protocol}\\
To evaluate the mechanical response of the integrated soft structure, both passive lattice and active actuator configurations were tested under mechanical and pneumatic loading conditions. Compression and bending tests were performed to assess deformation behavior, hysteresis, and actuation capabilities under cyclic conditions. Blocking force measurements were conducted to quantify the actuator’s output under constrained conditions.

Cubic lattice samples measuring 62.5 mm per side were fabricated using an IWP-type TPMS geometry with 12.5 mm unit cell size and 1.5 mm wall thickness. Each sample (N = 5) was tested using a Universal Testing Machine (Zwick/Roell, Germany) at a constant rate of 10 mm/min following a 1 N preload. Compression was applied to 20\%, 40\%, and 60\% strain to evaluate nonlinear stiffness and hysteresis.

Bladder-like actuators composed of six serially connected pneumatic cells (Figure \ref{supp-fig:actuator}, \ref{supp-fig:scalability}) were embedded halfway into matching lattice volumes (Figure \ref{supp-fig:halfembedactuator}). Actuators (N = 3) were driven by sinusoidal pressure inputs ranging from -20 kPa to 50 kPa. Bending deformation was recorded on video and analyzed using an image processing program (ImageJ \cite{abramoff2004image}). For blocking force measurements, the actuators were mounted on a micrometric translation stage (M-111.1DG, Physik Instrumente, Germany) interfaced with a triaxial load cell (ATI Nano 17, ATI Industrial Automation, United States). A half-wave rectified sine pressure input of up to 50 kPa was used to assess peak force generation at maximum extension. For free bending measurements, the actuators were mounted to an optical table (Standa, Lithuania) while actuators were driven with pressures from -20 kPa to 50 kPa in increments of 10. Sensor data was recorded over 100 cycles during both blocking force and free bending tests to assess repeatability (Figure \ref{supp-fig:cyclictest}). Images were taken (Nikon D7500, Nikon, Japan) and bending angles were analyzed in ImageJ.

To map the MELEGROS workspace boundaries, an electromagnetic probe (AURORA, NDigital, Canada) was placed at the tip of the dorsal finger (Figure \ref{fig:design}E) as movement swept all modes of reaching, pinching, and dorsal-ventral bending.

\threesubsection{Simulation of Integrated Lattice}\\
The simulation of MELEGROS was supported by a meshing workflow that converted CAD geometries into finite element models for SOFA simulations. STEP files defined the lattice volume, membranes, and cavities. In the sensorized configuration, additional STEP files specified the optical channels. All parts were meshed in Gmsh~\cite{geuzaine_gmsh_2009} through a Python interface. The workflow involved three stages: importing the geometry and removing duplicates, fragmenting overlapping volumes, and generating a tetrahedral mesh. The final meshes were exported in VTK format.

A subsequent procedure merged the separate component meshes into a single unstructured grid. This was achieved by detecting nodes within a specified tolerance and consolidating them, thereby ensuring a consistent connectivity across membranes, cavities, and lattice volume. The resulting merged mesh provided the mechanical basis for the simulator, where material properties were assigned according to homogenization tests.

Within the simulator, the homogenized properties of the lattice, identified from representative element compression tests, were assigned to the global volume. The pneumatic membranes were treated as deformable regions mapped to the volume, while the internal cavities defined pressure application surfaces. This arrangement enabled five independent pneumatic subsystems: three at the base, producing reaching and bending, and two at the tip, responsible for opening and closing. In the sensorized variant, the optical waveguides were introduced as elastic subdomains with higher stiffness than the surrounding lattice, allowing the positions of interest to be tracked during movement.

Control logic was implemented through dedicated scripts that read time-varying pressure inputs, applied them to the cavities, and monitored structural responses. Monitors tracked nodal displacements, contact interactions with rigid objects, and reaction forces in predefined regions of interest. The framework also allowed interactive actuation through manual commands, enabling exploration of different loading scenarios in real time. This integrated modeling framework allowed testing of actuation strategies and evaluation of the mechanical influence of the embedded sensors.

The detailed implementations and simulation results are provided in the Supplementary Material. Specifically, Algorithm~\ref{supp-alg:simulation} outlines the scene construction pipeline, while Algorithm~\ref{supp-alg:control} defines the pressure control logic. Two complementary models were realized: the \emph{Sensor-Position Model} (Model~A; Figure~\ref{supp-fig:sensor_position_model}), which specifies candidate sensor sites, and the \emph{Sensor-Integrated Model} (Model~B; Figure~\ref{supp-fig:sensor_integrated_model}), in which sensor bodies are mechanically embedded. To visualize sensor trajectories, Figure ~\ref{supp-fig:spm_open2}--\ref{supp-fig:spm_close2} report 2D projections for Model~A under the ventral-finger modes \textbf{open2} and \textbf{close2}, whereas Figure ~\ref{supp-fig:sim_open}--\ref{supp-fig:sim_close} illustrate the corresponding kinematics for Model~B under simultaneous-finger modes \textbf{open} and \textbf{close}. These graphs form the basis for Figure~\ref{fig:sensors}, linking sensor trajectories to the extracted angular variations. In addition, Figure~\ref{supp-fig:dissection} provides a broken-out view of the manipulator, highlighting the lattice, membranes, pneumatic cavities, sensors, and envelope considered in the simulations, while Figure~\ref{supp-fig:angle_sensors36} presents the angle--pressure relationships of two integrated tactile sensors, emphasizing the influence of sensor positioning on the measured angular response.

\threesubsection{Sensing Setup}\\
The photoemitters used in the devices (infrared LEDs, VSMY1850) have a peak emission at 850 nm, corresponding to the peak reception of the photoreceivers (VEMT7100X01). Signals were acquired with a custom PCB. Data collection was done with Python 3.13, and data analysis was performed using MATLAB (The MathWorks, Inc.).

\threesubsection{Design of Pneumatic Actuators and Control Setup}\\
To control the pneumatic actuators of MELEGROS, an I/O Device (NI USB-6218, National Instruments, United States) was used connected to a simple LabVIEW (National Instruments, United States) program. A dedicated analog output was attached to positive (ITV-0010, SMC, Japan) and negative (ITV-0090, SMC, Japan) pressure regulators to control the actuation pressure. The output of the regulators were attached to a 3/2 solenoid (V114A-5LOU, SMC, Japan) for each bladder. Each bladder was interfaced with a 2-port solenoid valve (VDW20GA, SMC, Japan) to maintain pressures while switching bladders. A dedicated pressure and vacuum (VCP 80/VCP 130, VWR, United States) was used during operation.

\medskip
\textbf{Supporting Information} \par 
Supporting Information is available from the authors.

\medskip
\textbf{Acknowledgements} \par 
This work was funded by the European Union Horizon 2020 research and innovation programme under grant agreement No. 863212 (PROBOSCIS project). The authors would like to thank Seonggun Joe (currently at the Korea Aerospace University) for the initial discussions on integrating pneumatic actuators in lattice structures when he was at the SBRP lab at IIT.
\medskip

\medskip

\textbf{Author Contributions} \par
Petr Trunin, Diana Cafiso, and Anderson Brazil Nardin contributed equally to this work.
Petr Trunin: Writing – review and editing, Writing – original draft, Visualization, Validation, Methodology, Investigation, Formal analysis, Data curation, Conceptualization.
Diana Cafiso: Writing – review and editing, Writing – original draft, Visualization, Validation, Methodology, Investigation, Formal analysis, Data curation, Conceptualization.
Anderson Brazil Nardin: Writing – review and editing, Writing – original draft, Visualization, Validation, Methodology, Investigation, Formal analysis, Data curation, Conceptualization.
Trevor Exley: Writing – review and editing, Writing – original draft, Visualization, Validation, Methodology, Investigation, Formal analysis, Data curation, Conceptualization.
Lucia Beccai: Writing – review and editing, Supervision, Resources, Project administration, Methodology, Funding acquisition, Conceptualization.
\medskip

\bibliographystyle{MSP}


\end{document}


\pagestyle{fancy}
\rhead{\includegraphics[width=2.5cm]{no-logo.png}}

\vspace{1cm}
\Large{Supporting Information}\\
\vspace{1cm}
\textbf{Monolithic Elephant-inspired Gripper with Optical Sensors}
\vspace{1cm}


\author{Petr Trunin$^\dagger$}
\author{Diana Cafiso$^\dagger$}
\author{Anderson Brazil Nardin$^\dagger$}
\author{Trevor Exley*}
\author{Lucia Beccai*}

\vspace{0.5cm}

\section*{Workflow}
\begin{figure}[h]
    \centering
    \includegraphics[width=0.9\linewidth]{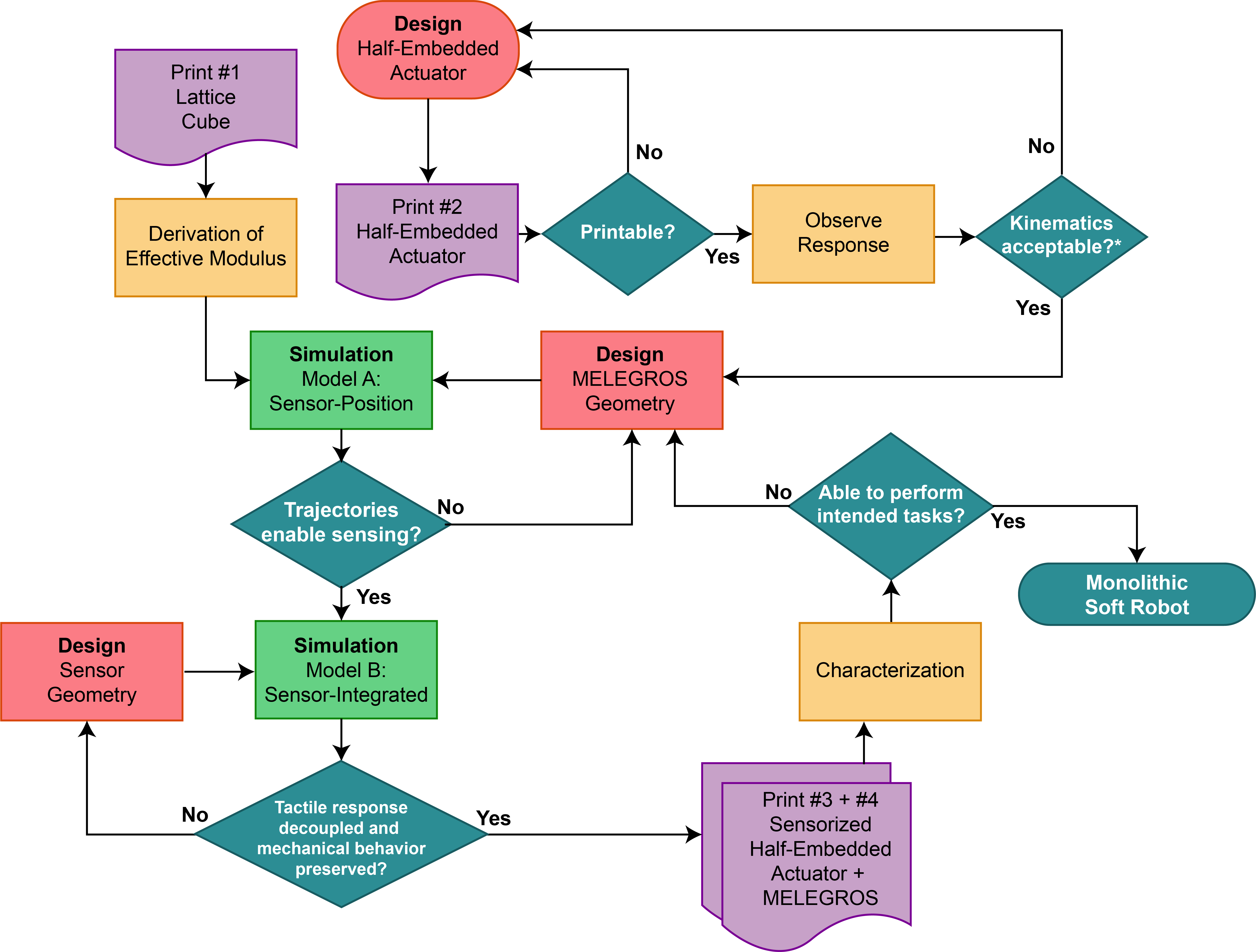}
    \caption{Workflow for MELEGROS development. *Observed kinematics should show that bending is representative to that of a finger to be deemed acceptable.}
    \label{fig:workflow}
\end{figure}

\justifying
Monolithic systems that integrate sensing and actuation into the fabrication process require specific workflows. As soft robotic simulation techniques advance, iterative design can be reinforced within the SOFA framework, as presented in Figure \ref{fig:workflow}. The process starts with the design: an actuation unit is printed (in this case, a half-embedded actuator) to assess printability and understand the kinematics during actuation. In parallel, the chosen lattice is printed to inform the homogenized envelope stiffness in the simulation. From here, the design of the gripper is simulated and observed to inform the sensor design. Depending on the sensor geometry, tactile and proprioceptive regions of interest are initially estimated. Representative positions of these regions are tracked during full range of movement (from -50 kPa to 50 kPa). If these observed trajectories produce (in the embedded sensors) the required bending amplitude so that a sensing response is elicited, then the sensors are physically added to the simulated gripper model. After minimal iterations, the same trajectories are collected to ensure that the addition of sensors does not change mechanical behavior (\textit{e.g.}, change in bending angle $<$ 5$^\circ$), and waveguide behavior is decoupled (\textit{e.g.}, tactile waveguide angle $<$ 5$^\circ$ during finger bending). From here, the design with integrated sensors is printed and tested for characterization and grasping applications. 
Only 4 different prints are required in the entire process: the actuation unit and lattice are printed first, and after sensor placement, a sensorized actuation unit for characterization and a gripper with integrated sensors are printed. This dramatically reduces multiple trial-and-error iterations. Closed-loop control is not addressed in this study and will be addressed in future work.

\clearpage

\section*{Lattice Pattern}

\begin{figure}[h]
    \centering
    \includegraphics[width=0.85\linewidth]{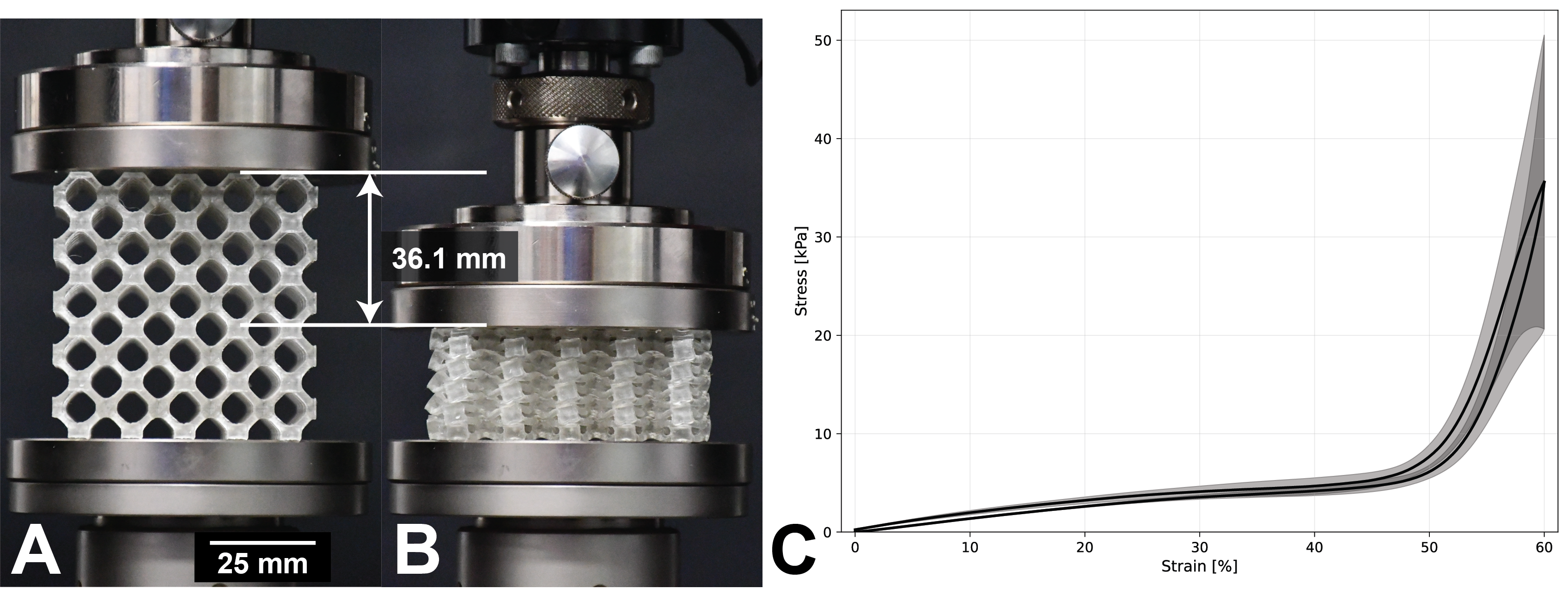}
    \caption{Lattice characterization test. Lattice in (A) undeformed and (B) deformed state. (C) Resulting stess-strain curve. Shaded areas represent $\pm$1 standard deviation (N = 5)}
    \label{fig:lattice}
\end{figure}

The IWP-type (Schoen I-graph Wrapped Package) TPMS (Triply Periodic Minimal Surface) lattice is defined by the following equation \cite{zhangVibrationCharacteristicsAdditive2024}:

\begin{multline}
\label{eq:TPMS}
    F(x,y,z) = 2(\cos(2\pi \frac{x}{L})\cos(2\pi \frac{y}{L}) + \cos(2\pi \frac{y}{L})\cos(2\pi \frac{z}{L}) + \cos(2\pi \frac{z}{L})\cos(2\pi \frac{x}{L}))\\
    -(\cos(4\pi \frac{x}{L}) + cos(4\pi \frac{y}{L}) + \cos(4\pi \frac{z}{L})) = t
\end{multline}
in which L is the unit cell size and t is the isovalue. In this study, the lattice was generated by 4D\_Additive Manufacturing Software Suite (Coretechnologie, Germany) with direct control over the unit cell size and minimum thickness of the struts. Cyclic compression tests were conducted on a cube of the lattice structure to inform the mechanical properties of the simulation (Figure \ref{fig:lattice}).

\section*{Actuator Design and Scalability}

\begin{figure}[hbt!]
    \centering
    \includegraphics[width=0.35\linewidth]{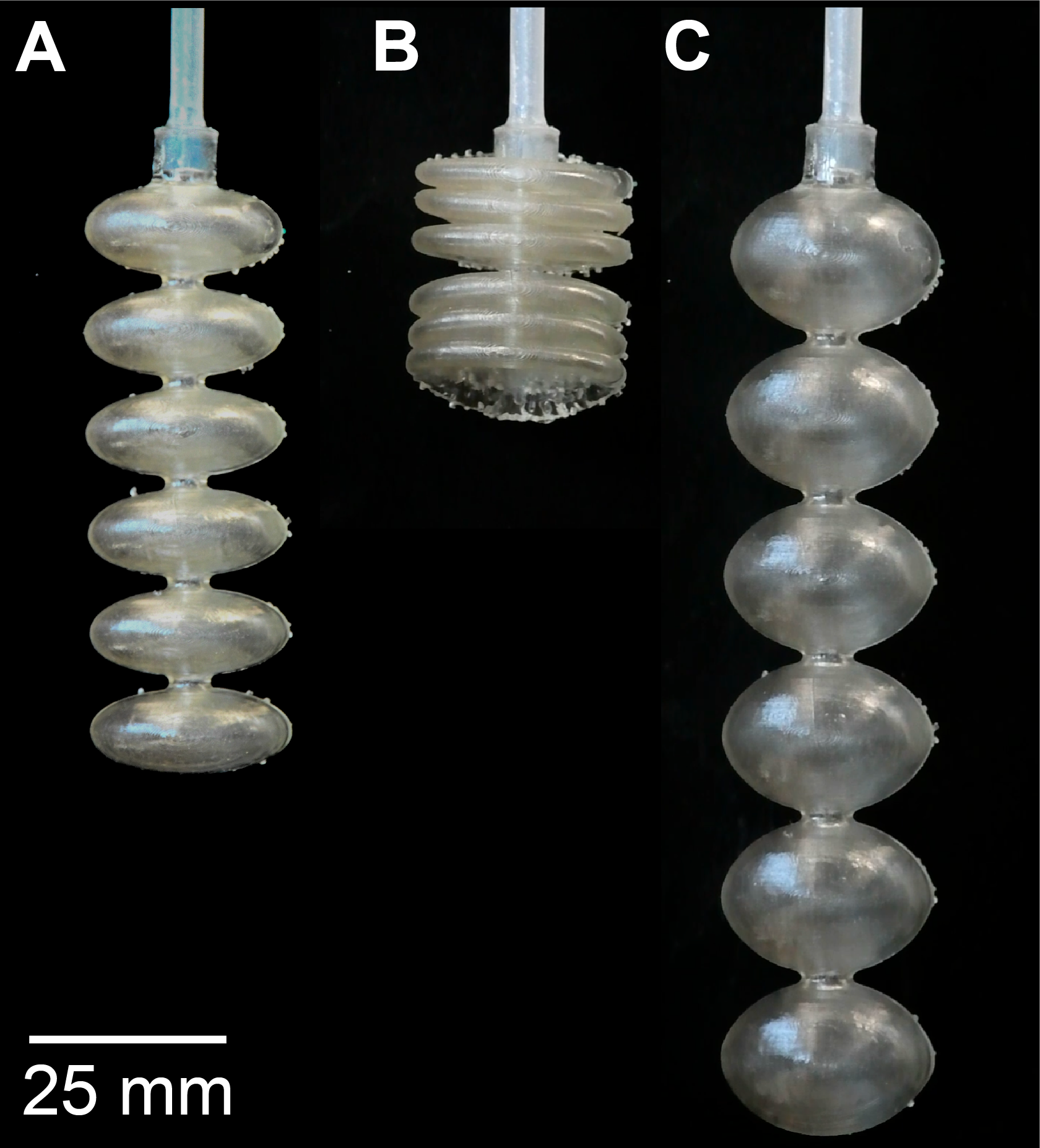}
    \caption{Bladder-like actuator (d = 25 mm) (A) at rest, (B) vacuum ($-50$ kPa), and (C) pressurized ($50$ kPa).}
    \label{fig:actuator}
\end{figure}

Bladder-like actuators were selected to target both compression and elongation in MELEGROS, under negative and positive pressure, respectively, with negligible radial expansion (Figure \ref{fig:actuator}). 

\begin{figure}[hbt!]
    \centering
    \includegraphics[width=0.5\linewidth]{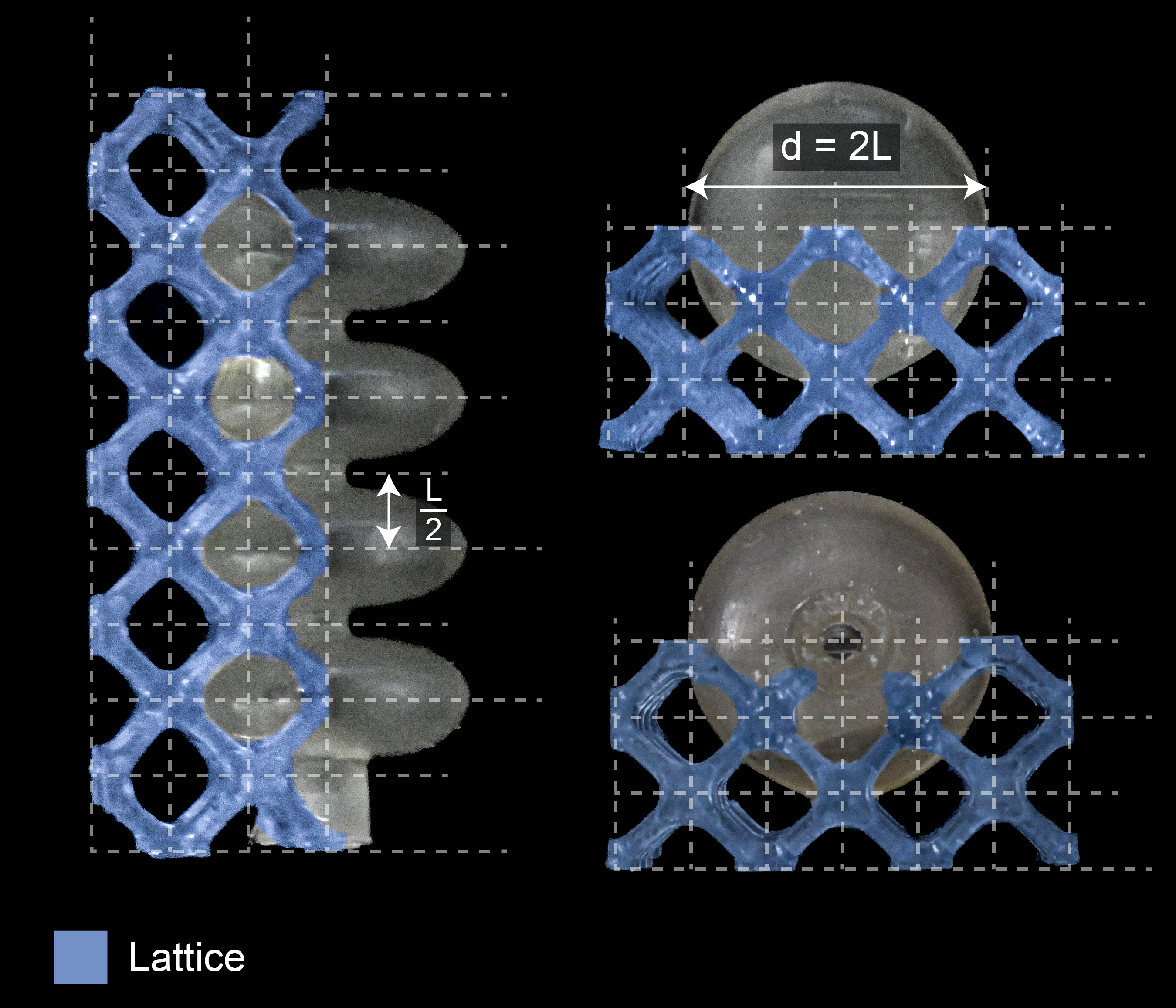}
    \caption{Spacing of the IWP-type TPMS lattice with respect to the bladder when in half-embedded configuration. Bladder dimensions such as d, diameter, are dependent on L, unit cell size.}
    \label{fig:dimensions}
\end{figure}

The dimensions of these actuators are directly related to the connections to the IWP-type TPMS lattice. Thus, it is important to design the bladder diameter and length accordingly. For example, given a unit cell size of $L$, the separation between bladders should equal $L$, and the diameter of each bladder should equal $2L$ (Figure \ref{fig:dimensions}).

\begin{figure}[hbt!]
    \centering
    \includegraphics[width=0.75\linewidth]{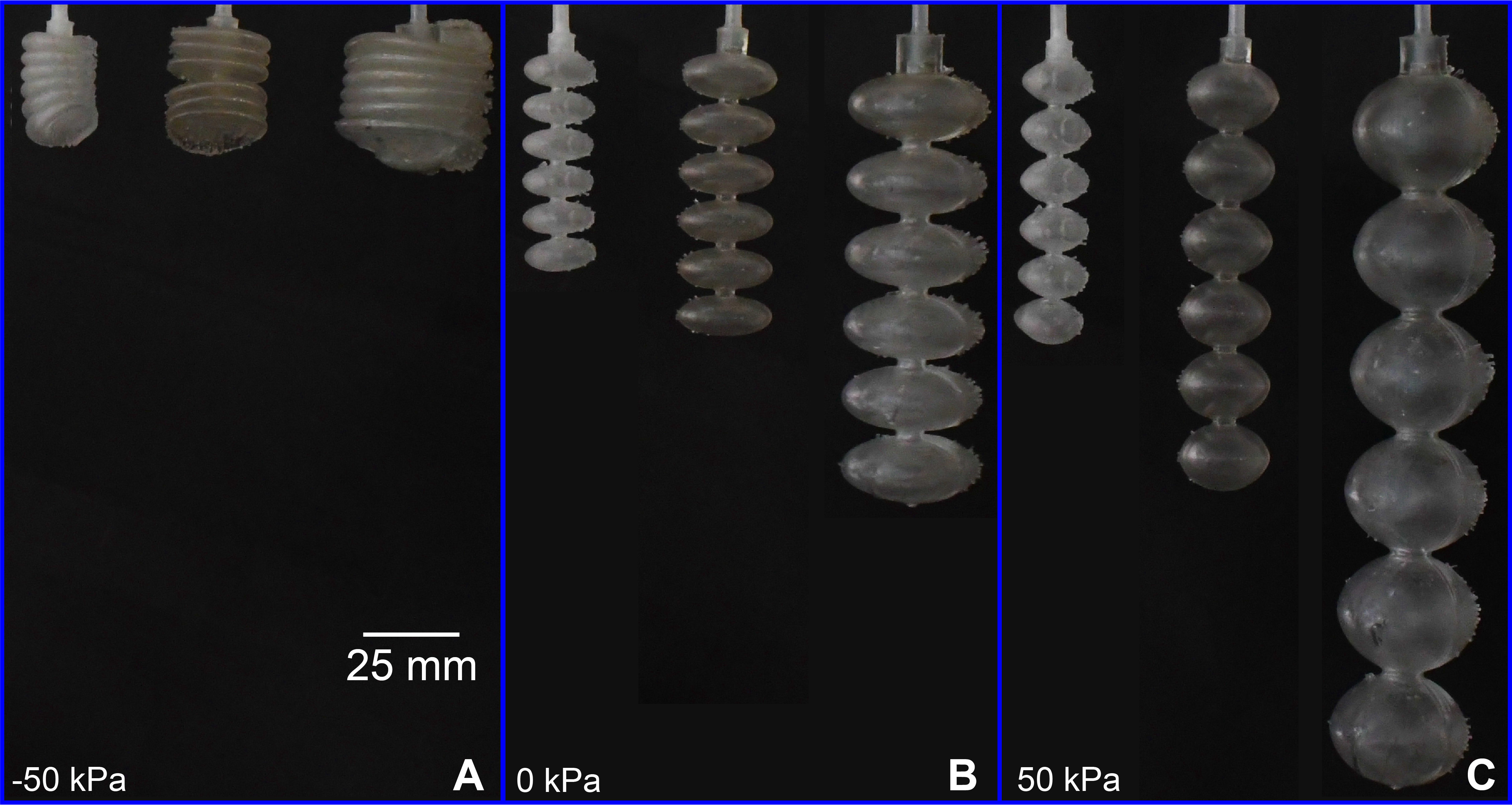}
    \caption{The bladder-like actuators (d = 25 mm) scaled 0.75, 1.0, and 1.25 in the (A) compressed, (B) resting, and (C) actuated states.}
    \label{fig:scalability}
\end{figure}

\begin{figure}[hbt!]
    \centering
    \includegraphics[width=0.5\linewidth]{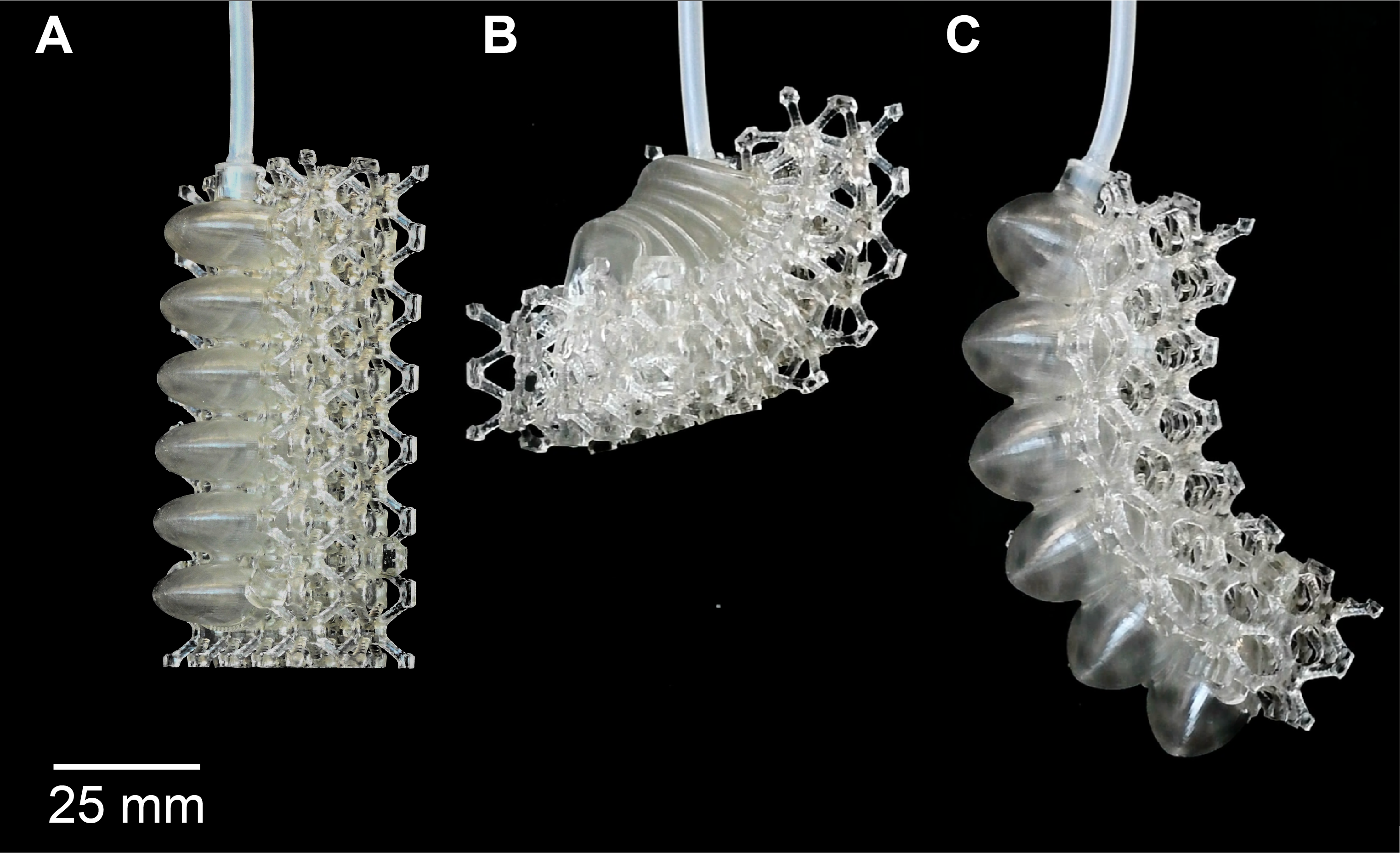}
    \caption{Half-embedded bladder-like actuator (A) at rest, (B) vacuum ($-50$ kPa), and (C) pressurized ($50$ kPa).}
    \label{fig:halfembedactuator}
\end{figure}

Preliminary scaling studies confirmed that both lattice and actuators can be uniformly enlarged or reduced without altering the fabrication workflow, however smaller chambers prove more difficult to remove uncured material during post processing (Figure \ref{fig:scalability}). The strain-limiting behavior was achieved by embedding actuators halfway into the lattice (Figure \ref{fig:halfembedactuator}): radial expansion was constrained while bending was directed along the gripper’s axis. In addition, the lattice provided support during the 3D-printing process, ensuring stable fabrication of the thin-walled bladders and complex geometries. Since the lattice and actuator size are dependent on each other, the smaller the actuator also significantly increases the amount of required material for the lattice. 

\clearpage

\section*{Gripper Design}

\begin{figure}[hbt!]
    \centering
    \includegraphics[width=0.9\linewidth]{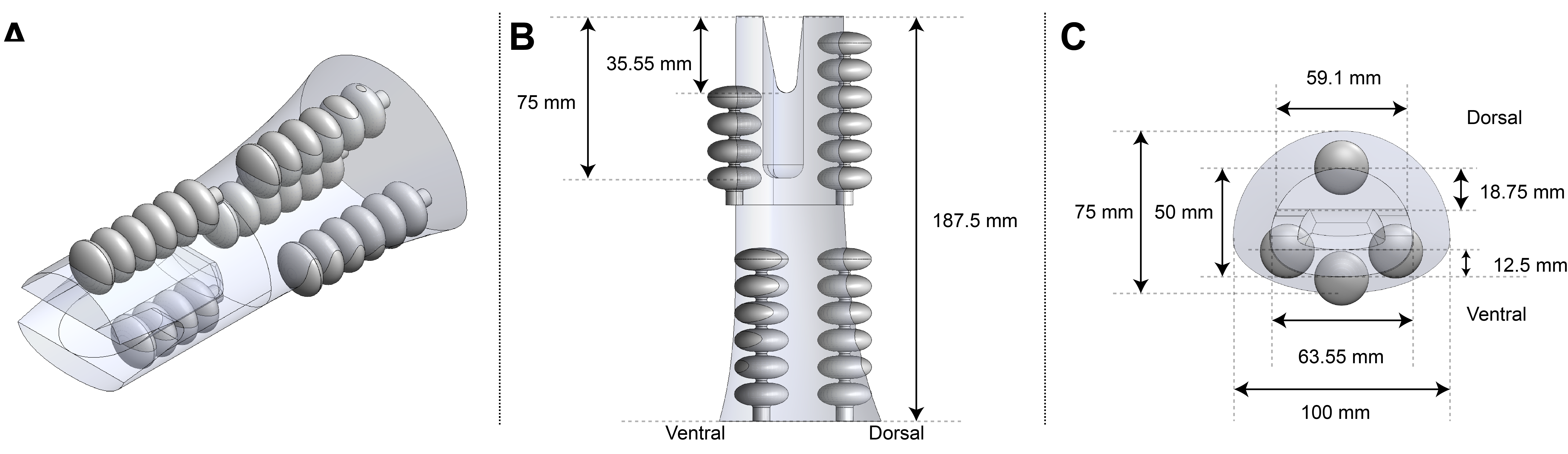}
    \caption{Design of MELEGROS (A) isometric view, (B) frontal view, and (C) top view, with dimensions. The envelope used to generate the lattice is shown in transparency.}
    \label{fig:melegrosdimensions}
\end{figure}

The final design of MELEGROS is shown in Figure \ref{fig:melegrosdimensions} without sensor placements. The transparent envelope (bulk volume) is input to 4D\_Additive Manufacturing Software Suite to generate the lattice body. The base and finger actuators are located with respect to the lattice as Figure \ref{fig:dimensions} depicts. The fingers are designed asymmetrically, with varying geometry (width, profile, thickness) and number of actuator chambers detailed in Figure \ref{fig:melegrosdimensions}B,C. The dorsal actuator spans the entire finger length and drives 'active' movement, while the ventral actuator covers only a fraction of the finger, leaving the distal part 'passive'. This asymmetry produces more behaviorally bioinspired motions, such as scooping.

\clearpage

\section*{Sensor Design}
FEM simulations were performed in COMSOL Multiphysics\textsuperscript{\textregistered} (COMSOL Inc., Sweden) using a 2D approximation to reduce computational cost. The Ray Optics and Solid Mechanics interfaces were coupled to capture deformation-dependent optical loss. The waveguide was modeled as a rectangular domain with an emitter at one end and a Ray Counter at the opposite end. 

\begin{figure}[hbt!]
    \centering
    \includegraphics[width=0.75\linewidth]{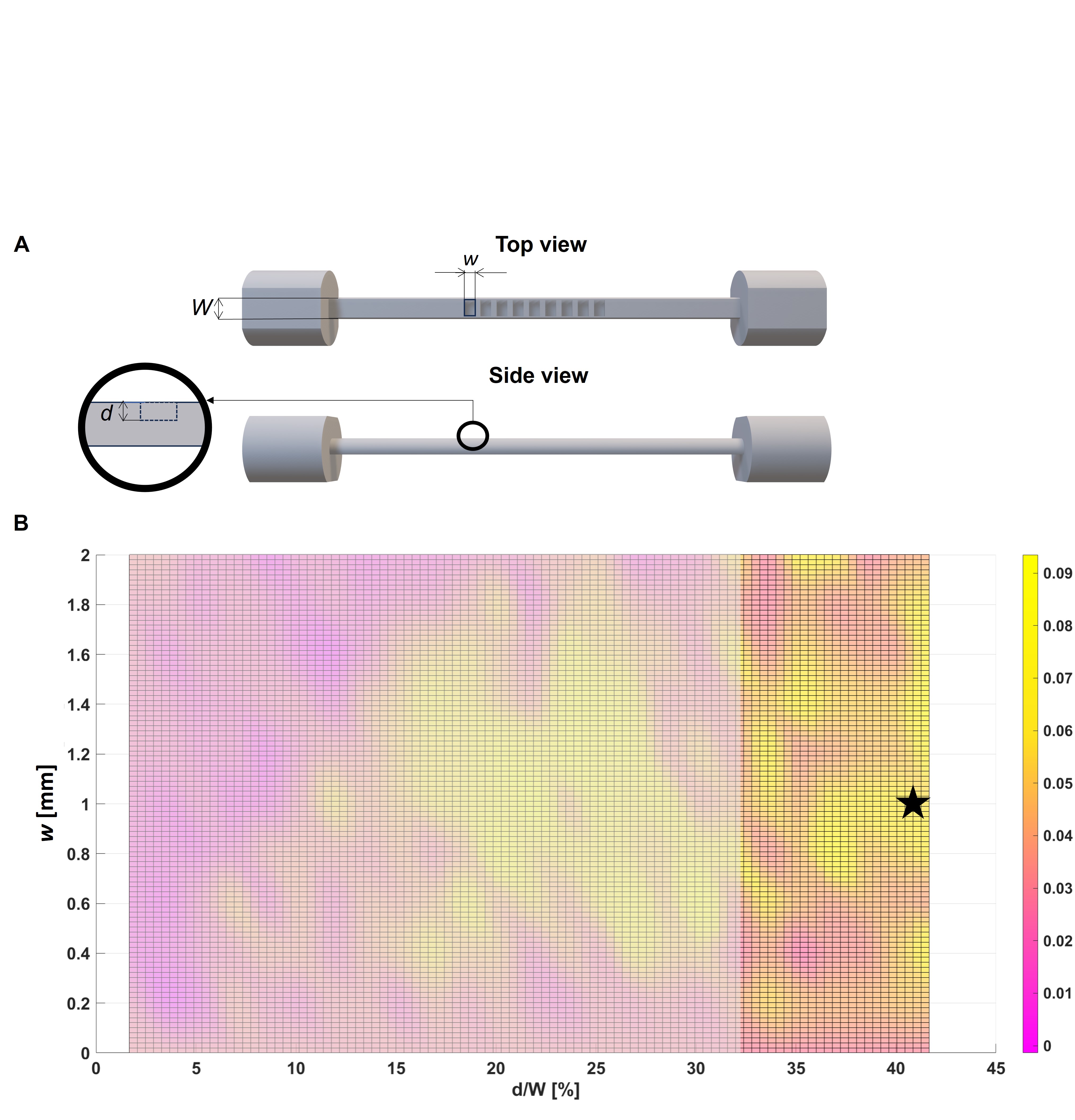}
    \caption{(A) Waveguides design choice based on the printability and (B) simulation in COMSOL with scale bar for Goodness Coefficient..}
    \label{fig:sensingdesign}
\end{figure}

To identify geometries that maximize sensitivity while preserving linearity, we defined a Goodness Coefficient (GC) as the ratio of the optical intensity drop ($\Delta$I) to the RMSE of a linear regression. The bigger the GC, the better the sensing performance is expected. Two main parameters were used in the simulation: the width of the superficial pattern, w, and the ratio between the depth of the pattern, d, and the width of the waveguide, W. There were 25 parametric steps for d and 20 for w. Furthermore, the size of the electrical components (1 mm) was considered when defining the depth of the pattern (the bottom of the well must not coincide with the photoemitter by default). The simulation was explained further in Trunin et al. \cite{truninDesign3DPrinting2025b}(Figure \ref{fig:sensingdesign}).

\begin{figure}[hbt!]
    \centering
    \includegraphics[width=0.85\linewidth]{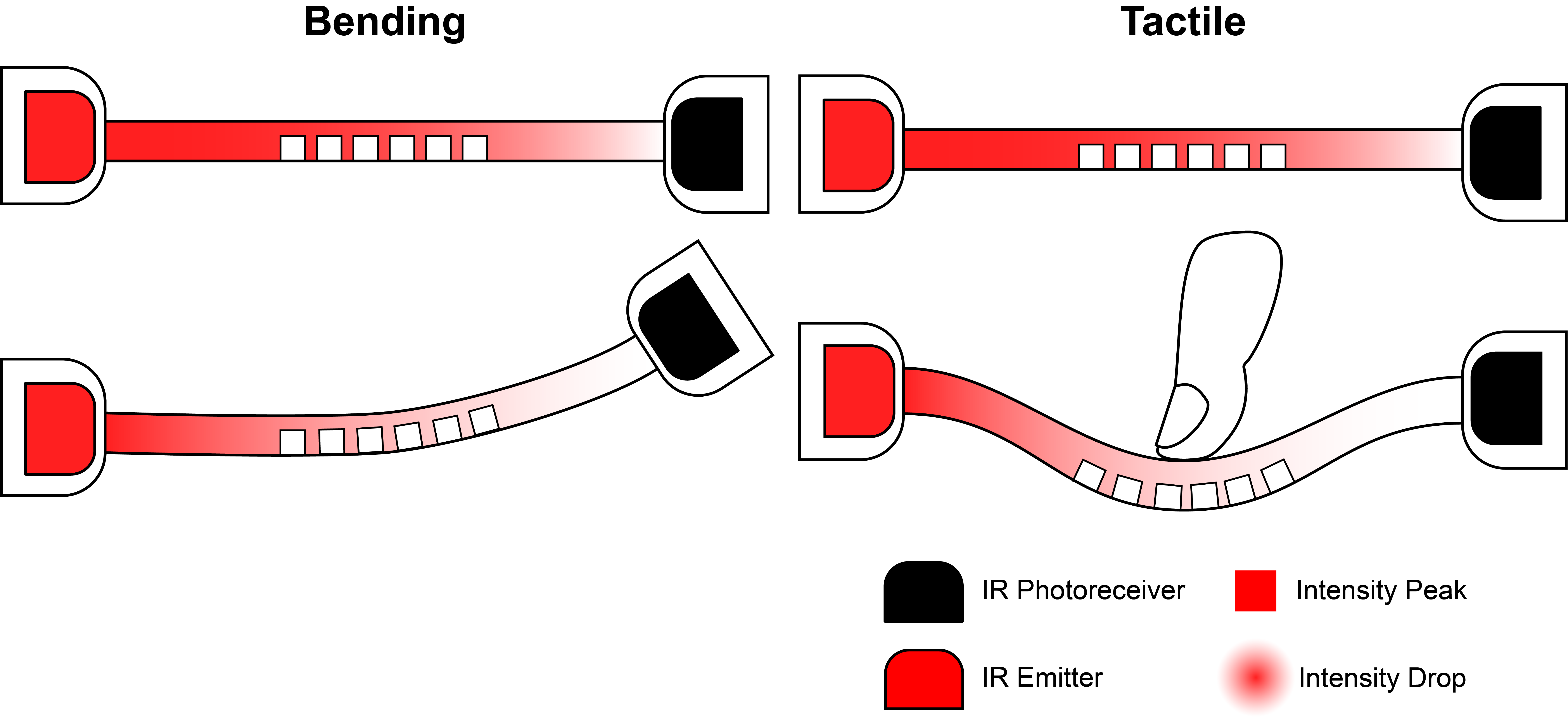}
    \caption{Working principle of bending and tactile sensors. Light intensity drops after deformation in the planned direction. }
    \label{fig:sensorbehavior}
\end{figure}

Both proprioception and tactile sensors are designed and integrated in the lattice structure with superficial wells on the side of planned deformation(bending). The tactile sensor bends due to external stimuli, while the proprioceptive sensor bends due to the structural bend (Figure \ref{fig:sensorbehavior}).

\section*{Cyclic test}
\begin{figure}[hbt!]
    \centering
    \includegraphics[width=0.85\linewidth]{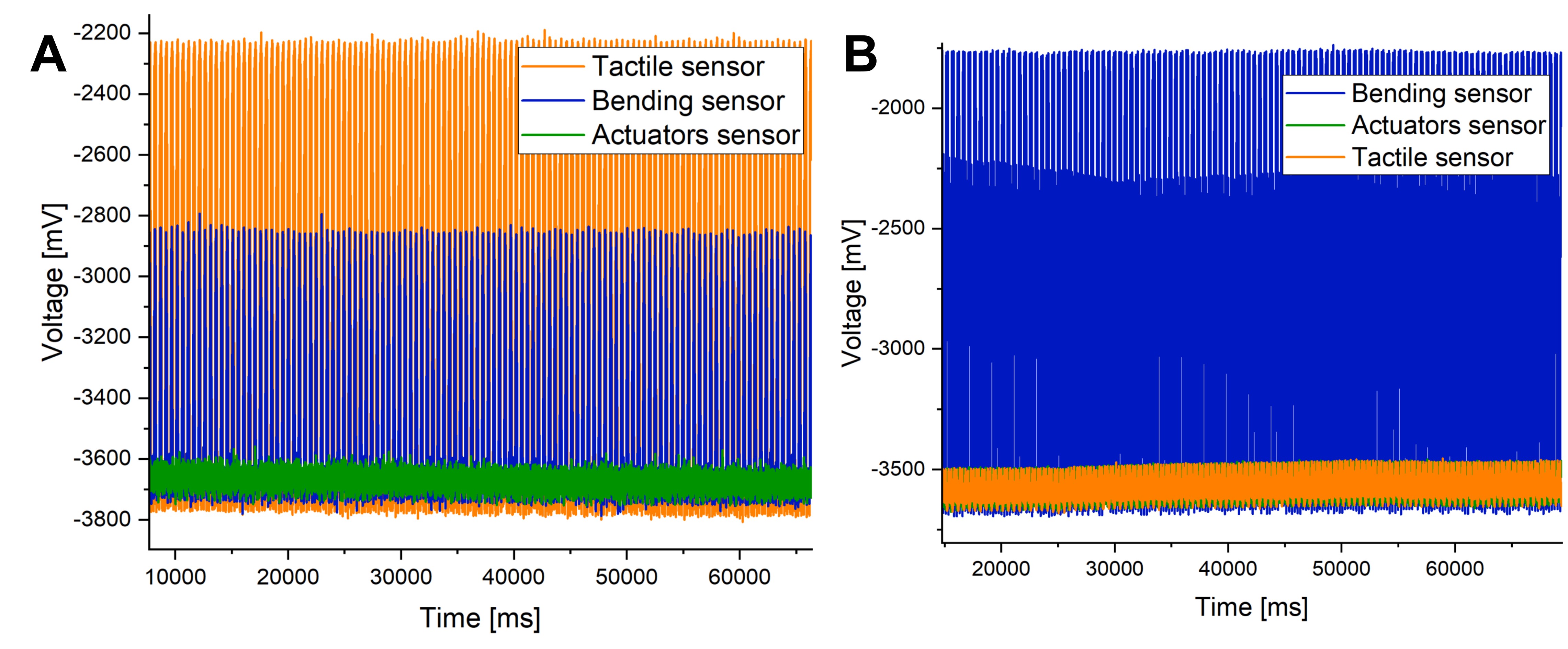}
    \caption{Sensor data from the tactile, bending, and actuator sensors during the cyclic test of the bladder-like actuator from (A) blocking force and (B) free bending tests.}
    \label{fig:cyclictest}
\end{figure}

Sensor data were recorded during cyclic tests of the actuator. For the blocking force test, 100 cycles of positive (50 kPa) pressure were applied. For the free-bending test, 100 cycles of positive (50 kPa) and negative (-20 kPa) pressures were applied (Figure \ref{fig:cyclictest}).

\clearpage

\section*{Simulation Framework}

To evaluate actuator kinematics and guide sensing, two simulation cases were implemented in SOFA: the \emph{Sensor-Position Model} (Model~A), in which only candidate sensor locations are indicated, and the \emph{Sensor-Integrated Model} (Model~B), in which the sensor bodies are also present and therefore contribute mechanically.

\begin{figure}[H]
    \centering
    \includegraphics[width=0.8\linewidth]{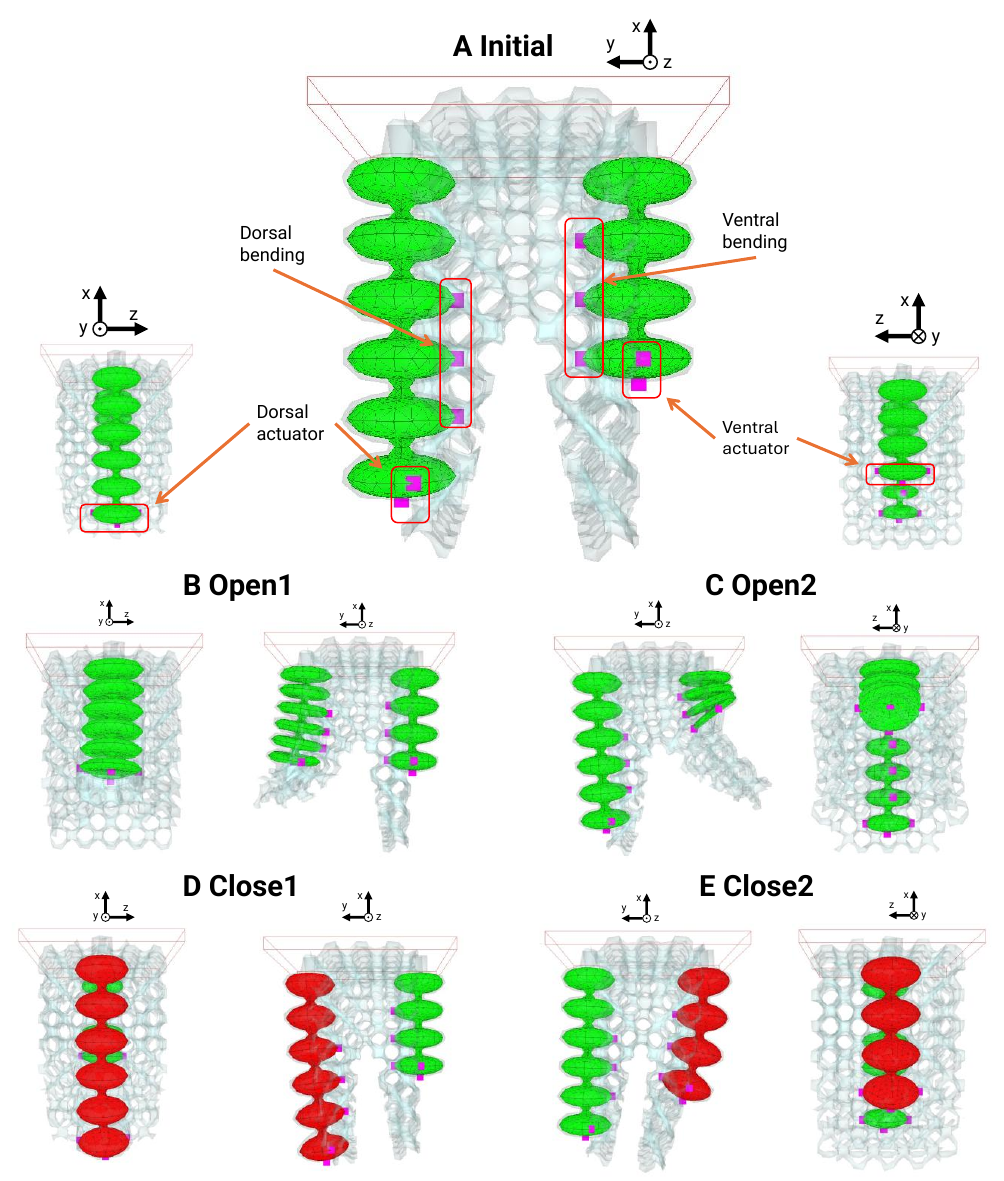}
    \caption{SOFA simulation setup for the \emph{Sensor-Position Model} (Model~A).  
    A: Initial configuration with candidate sensor locations highlighted.  
    B: Finger \textbf{open1} (dorsal finger).  
    C: Finger \textbf{open2} (ventral finger).  
    D: Finger \textbf{close1} (dorsal finger).  
    E: Finger \textbf{close2} (ventral finger). Green indicates zero to negative pressure, red indicates positive pressure, and pink marks the sensor positions of interest. }
    \label{fig:sensor_position_model}
\end{figure}

Figure~\ref{fig:sensor_position_model} summarizes Model~A. Figure~\ref{fig:sensor_position_model}A shows the initial configuration with candidate sensor locations highlighted. Figure \ref{fig:sensor_position_model}B–E illustrate finger-specific actuation modes: \textbf{open1} and \textbf{close1} for the dorsal finger, and \textbf{open2} and \textbf{close2} for the ventral finger.

\begin{figure}[H]
    \centering
    \includegraphics[width=0.8\linewidth]{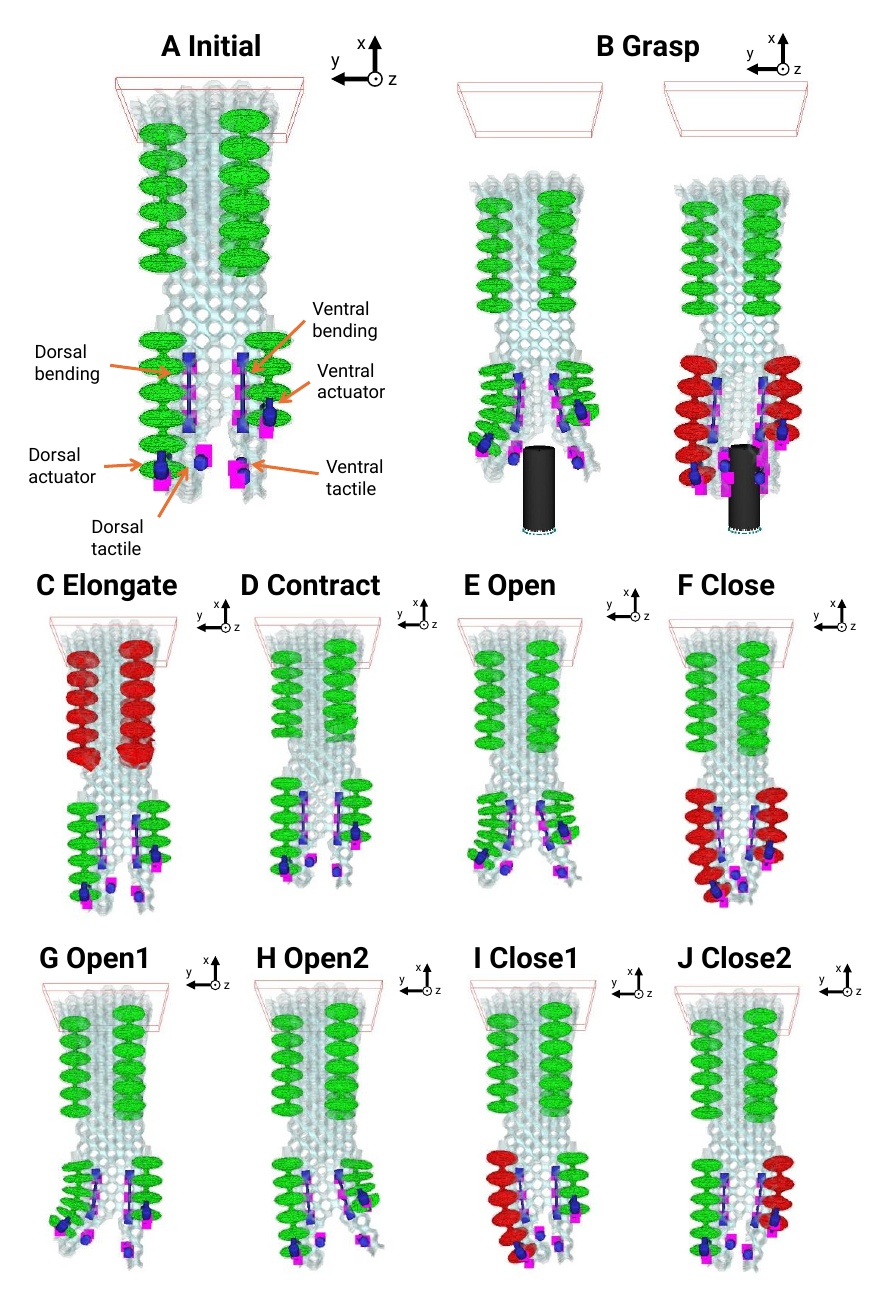}
    \caption{SOFA simulation setup for the \emph{Sensor-Integrated Model} (Model~B).  
    A: Initial configuration with integrated sensors.  
    B: \textbf{Grasp}.  
    C: \textbf{Elongate} proximal chambers.  
    D: \textbf{Contract} proximal chambers.  
    E: \textbf{Open} (both fingers simultaneously).  
    F: \textbf{Close} (both fingers simultaneously).  
    G: \textbf{open1} (dorsal finger).  
    H: \textbf{open2} (ventral finger).  
    I: \textbf{close1} (dorsal finger).  
    J: \textbf{close2} (ventral finger). Green indicates zero to negative pressure, red indicates positive pressure, pink marks monitored points, and blue marks the sensors that are integrated. }
    \label{fig:sensor_integrated_model}
\end{figure}

Figure~\ref{fig:sensor_integrated_model} reports Model~B, where sensors are embedded in the mechanics. Figure~\ref{fig:sensor_integrated_model}A shows the initial state; Figure~\ref{fig:sensor_integrated_model}B depicts the \textbf{grasp} sequence. Figure \ref{fig:sensor_integrated_model}C–D illustrate \textbf{elongate} and \textbf{contract} of the proximal chambers. The sensing layout comprises four sensor regions, namely \emph{Dorsal actuator}, \emph{Dorsal bending}, \emph{Ventral actuator}, and \emph{Ventral bending}.

Figure \ref{fig:sensor_integrated_model}E–J present additional finger actuation modes: \textbf{open} and \textbf{close} involve simultaneous motion of both fingers, whereas \textbf{open1}/\textbf{close1} correspond to the dorsal finger and \textbf{open2}/\textbf{close2} to the ventral finger. The integrated sensing layout comprises six sensors, namely \emph{Dorsal actuator}, \emph{Dorsal bending}, \emph{Dorsal tactile}, \emph{Ventral actuator}, \emph{Ventral bending}, and \emph{Ventral tactile}.

To reproduce the simulations, it is provided two reference algorithms. The first algorithm (Algorithm~\ref{alg:simulation}) focuses on creating the simulation environment, defining the soft arm’s physical parameters, importing the meshed subdomains (lattice/actuator volume, membranes, cavities, and, when applicable, sensor inclusions), and linking all models, ROIs, collision layers, and visualization. The second algorithm (Algorithm~\ref{alg:control}) provides the control logic that applies time-varying pressure inputs according to predefined regimes (the different modes of operation), while tracking motion via monitored points, logging reaction forces in a region of interest, and enabling manual teleoperation.

\begin{algorithm}[H]
  \caption{Scene Construction}
  \label{alg:simulation}
  \begin{algorithmic}[1]
    \State Initialize SOFA core (animation loop, solvers, collision pipeline, visualization); set $dt$ and gravity
    \State Add environment plane (mesh $\rightarrow$ topology $\rightarrow$ collision $\rightarrow$ visual)
    \State Create rigid target (pose, mass/inertia); add collision shell and visual mapping
    \State Define list \textit{Actuators} with pose parameters (rotation, translation)
    \For{\textbf{each} actuator $\in$ \textit{Actuators}}
        \State \textbf{Soft body (tet mesh)}
        \State \hspace{0.8em} Load merged tetrahedral mesh; create topology and mechanical state
        \State \hspace{0.8em} Fix base using \textit{BoxROI} + \textit{RestShapeSprings}; define distal ROI for force readout
        \State \hspace{0.8em} \textbf{Monitors:} \For{$j=1$ \textbf{to} $6$}
            \State \hspace{1.6em} Attach \textit{Monitor} to a triplet of node indices; enable position export
          \EndFor
        \State \textbf{Membrane subdomains}
        \State \hspace{0.8em} Gather STL files: \texttt{membrane*.stl}
        \For{\textbf{each} membrane file}
            \State \hspace{1.6em} Build \textit{MeshROI} on host tets; add dedicated FEM block
        \EndFor
        \State \textbf{Sensor subdomains (Model B)}
        \State \hspace{0.8em} Gather STL files: \texttt{sensor*.stl}
        \For{\textbf{each} sensor file}
            \State \hspace{1.6em} Build \textit{MeshROI} on host tets; add elastic waveguide block
        \EndFor
        \State \textbf{Pneumatic cavities}
        \State \hspace{0.8em} Gather STL files: \texttt{cavity*.stl}
        \For{\textbf{each} cavity file}
            \State \hspace{1.6em} Load surface; add \textit{SurfacePressureConstraint} (pressure value-type) with barycentric mapping
        \EndFor
        \State \textbf{Collision layer}
        \State \hspace{0.8em} Load outer surface; add triangle/line/point collision models and mapping
        \State \textbf{Visualization}
        \State \hspace{0.8em} Overlay translucent lattice; for $j=1,\dots,6$ overlay colored sensor visuals; map to tets
    \EndFor
    \State Attach controller (options: save on/off, automatic on/off, selected input regime); \textbf{return} root node
  \end{algorithmic}
\end{algorithm}

\begin{algorithm}[H]
  \caption{Pressure Controller}
  \label{alg:control}
  \begin{algorithmic}[1]
    \State Discover cavities in the scene; cache handles to their \textit{SurfacePressureConstraint}s
    \If{automatic mode}
      \State Load CSV profile for selected regime (open, close, open1, open2, close1, close2, elongate, contract, grasp)
      \State Build per-cavity sequences; set pre/post holds; compute total duration
    \EndIf
    \If{saving enabled} \State Open output CSV \EndIf
    \While{simulation running}
      \If{automatic window active}
        \For{\textbf{each} cavity}
          \State Set pressure value from time-indexed sequence
        \EndFor
      \EndIf
      \State Read back pressures; sample forces in ROI; compute $(F_x,F_y,F_z)$ and selected distances
      \If{saving enabled and past pre-hold} \State Append timestamp and pressures to CSV \EndIf
      \If{automatic mode and time $>$ total duration} \State Stop simulation \EndIf
      \State \textit{Teleoperation (manual mode):}
            number keys $\rightarrow$ adjust cavity pressures;
            arrows $\rightarrow$ translate rest pose; ``.''/``/'' $\rightarrow$ rotate rest pose
    \EndWhile
  \end{algorithmic}
\end{algorithm}

\begin{figure}[H]
    \centering
    \includegraphics[width=0.99\linewidth]{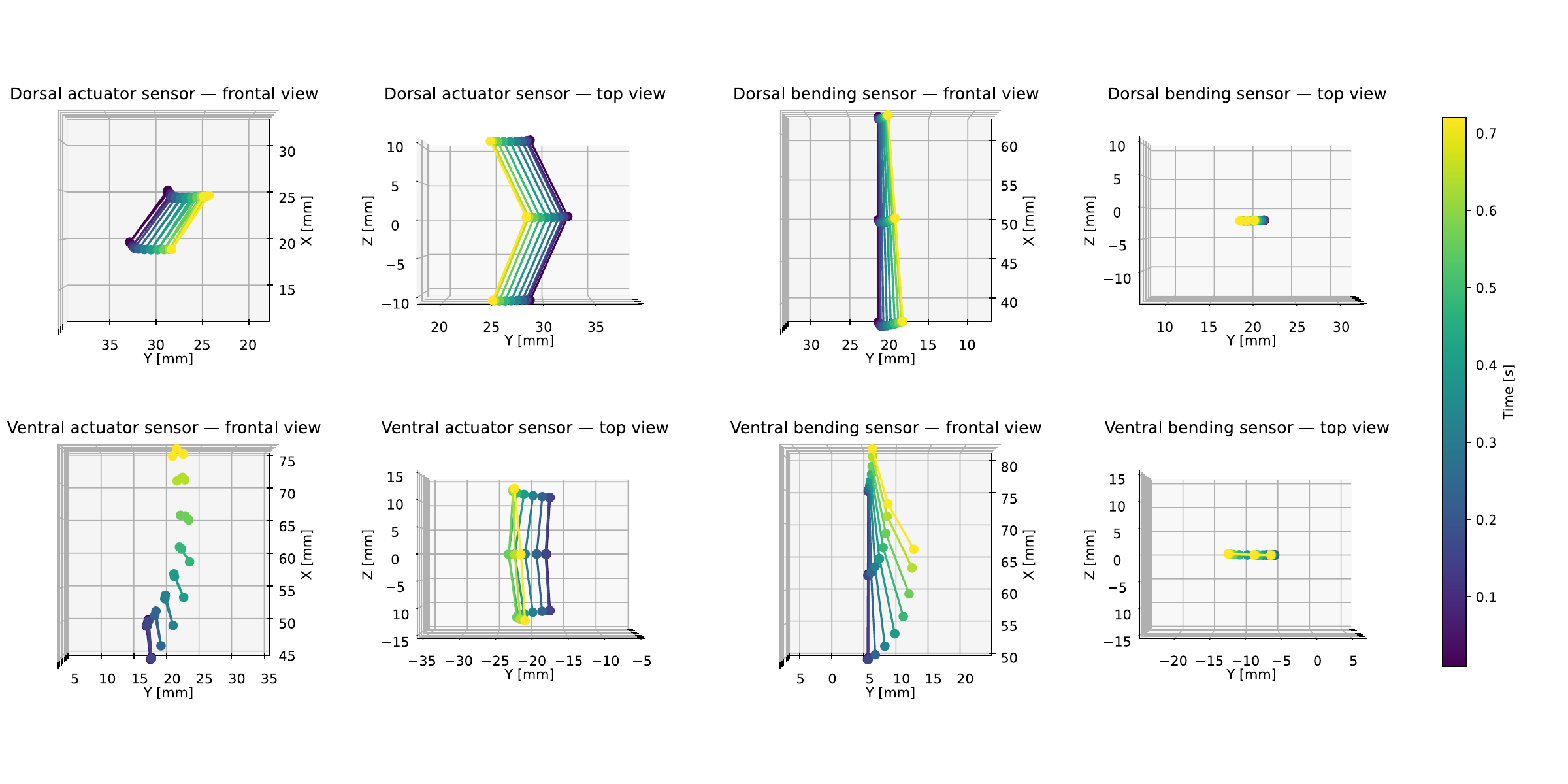}
    \caption{2D projections of sensor trajectories for the \emph{Sensor-Position Model} (Model~A).  
    Finger \textbf{open2}.}
    \label{fig:spm_open2}
\end{figure}

\begin{figure}[H]
    \centering
    \includegraphics[width=0.99\linewidth]{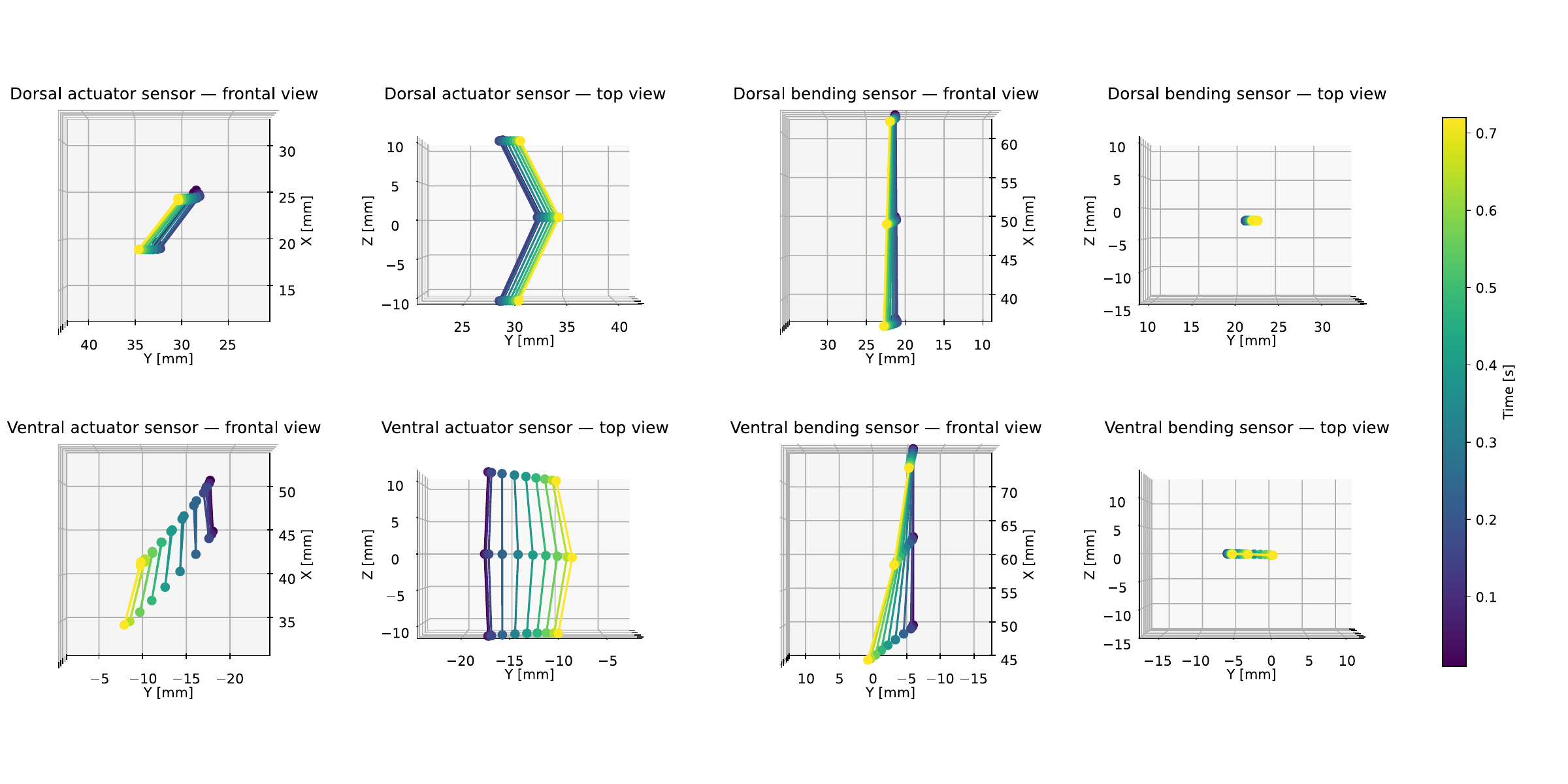}
    \caption{2D projections of sensor trajectories for the \emph{Sensor-Position Model} (Model~A).  
    Finger \textbf{close2}.}
    \label{fig:spm_close2}
\end{figure}

The 2D projections of the sensor-point trajectories for Model~A are provided under ventral-finger modes. These plots complement Figure~\ref{fig:sensor_position_model} by making the spatiotemporal paths explicit: \textbf{open2} in Figure~\ref{fig:spm_open2} and \textbf{close2} in Figure~\ref{fig:spm_close2}.

\begin{figure}[H]
    \centering
    \includegraphics[width=0.95\linewidth]{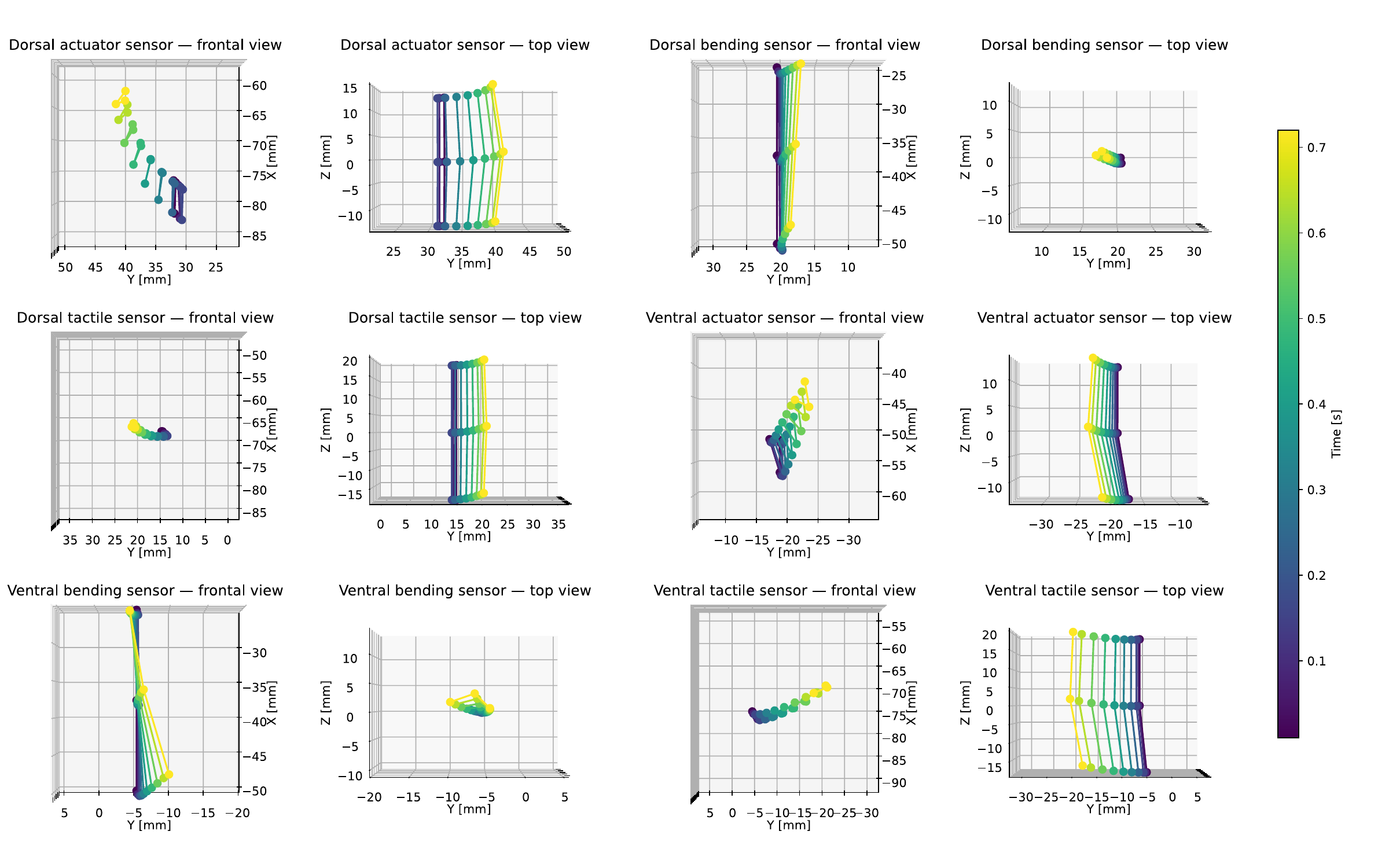}
    \caption{2D projections of sensor trajectories for the \emph{Sensor-Integrated Model} (Model~B). 
    Fingers \textbf{open}.}
    \label{fig:sim_open}
\end{figure}

\begin{figure}[H]
    \centering
    \includegraphics[width=0.95\linewidth]{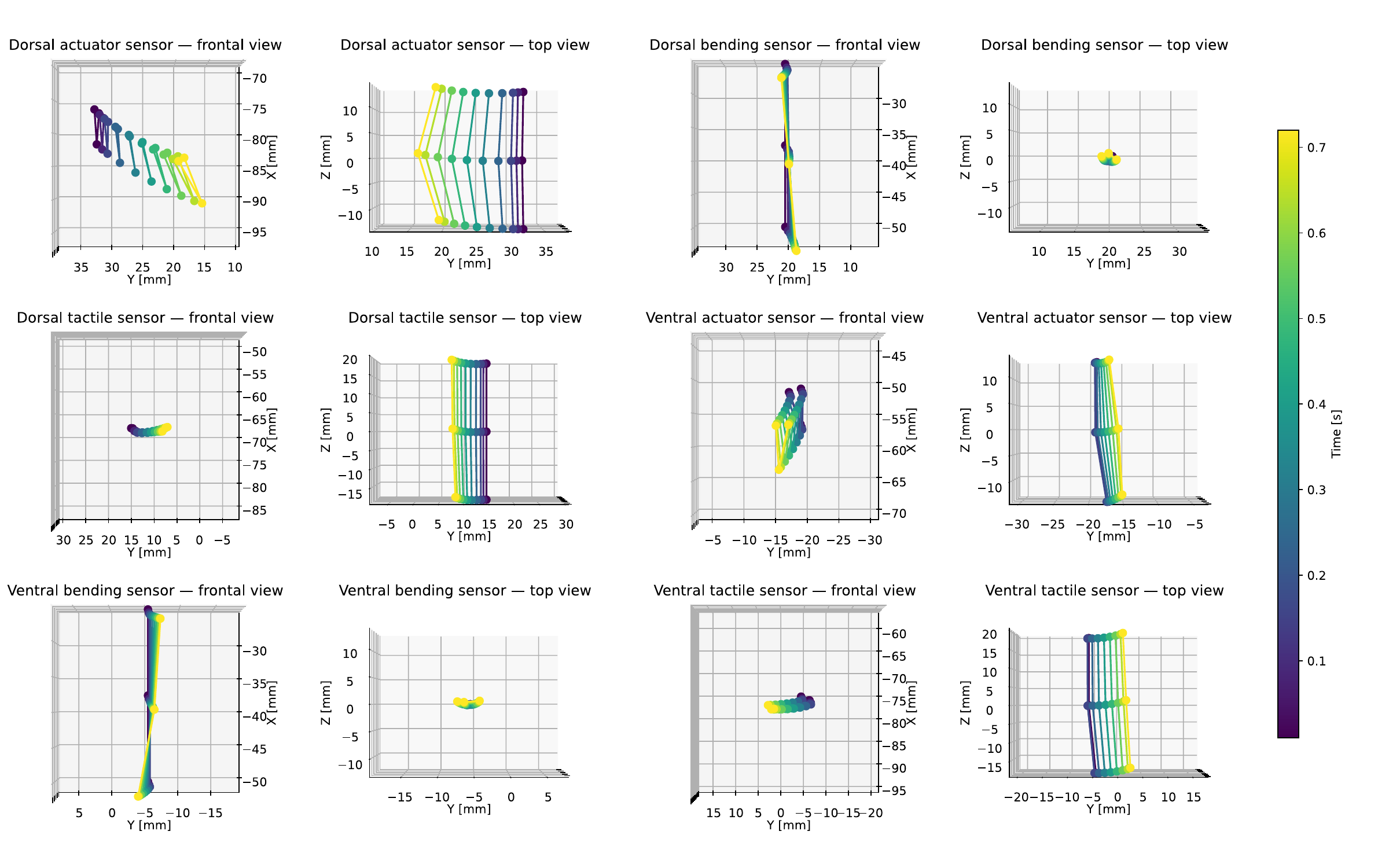}
    \caption{2D projections of sensor trajectories for the \emph{Sensor-Integrated Model} (Model~B). 
    Fingers \textbf{close}.}
    \label{fig:sim_close}
\end{figure}

Then, it is reported the corresponding projections for Model~B. These results visualize the effect of the integrated sensors on the kinematics across the finger-simultaneous modes: \textbf{open} (Figure~\ref{fig:sim_open}) and \textbf{close} (Figure~\ref{fig:sim_close}).

In Model~A (Figure \ref{fig:spm_open2} and \ref{fig:spm_close2}), the frontal-view trajectories provide clear evidence of the suitability of the site. The \textbf{open2} and \textbf{close2} modes correspond to ventral-finger actuation, where the finger is driven into opening and closing, respectively. In the ventral actuator sensor during the \textbf{open2} mode, the trajectory extends over roughly 20–30~mm, capturing large-scale motion of the finger. Under the \textbf{close2} mode, the same site still spans more than 15~mm, confirming its sensitivity across bi-directional actuation regimes. By comparison, the ventral bending sensor shows trajectories confined to narrow bands of about 5–10~mm in both modes, reflecting localized but consistent deformation. The nature of these responses supports the adoption of both sensor types for integration in Model~B. In Model~B (Figure \ref{fig:sim_open} and \ref{fig:sim_close}), the \textbf{open} and \textbf{close} modes correspond to the simultaneous actuation of both fingers. The actuator sensors trace excursions of about 20–25~mm in the \textbf{open} mode and remain around 15~mm in the \textbf{close} mode, showing that the integration of sensor bodies preserves their ability to capture large-scale motion across bi-directional actuation regimes. Bending sensors follow more confined trajectories in the 5–10~mm range, consistent with localized deformation signals identified in Model~A. Dorsal and ventral sites remain well separated, with smooth, distinct traces that reflect complementary roles in encoding overall motion and local bending.

\begin{figure}[H]
    \centering
    \includegraphics[width=0.95\linewidth]{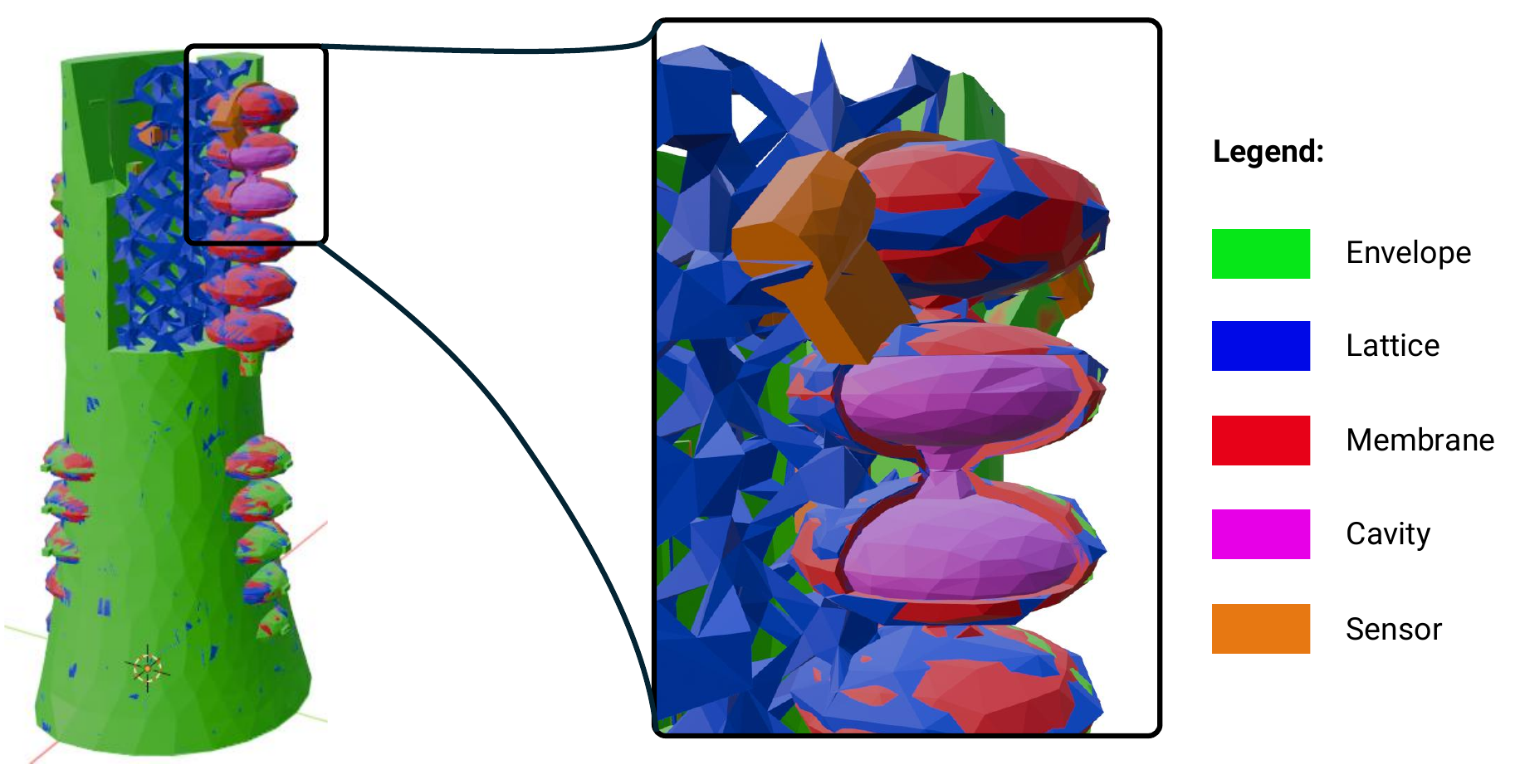}
    \caption{Broken-out section of the soft manipulator highlighting the simulation-relevant regions.}
    \label{fig:dissection}
\end{figure}

Figure~\ref{fig:dissection} illustrates a schematic of the modeled MELEGROS, highlighting the distinct components considered in the simulation process. The homogenized outer envelope is shown in green, the internal lattice structure in blue, the actuator membranes in red, the pneumatic cavities in magenta, and the embedded sensors in orange.

\begin{figure}[H]
    \centering
    \includegraphics[width=0.7\linewidth]{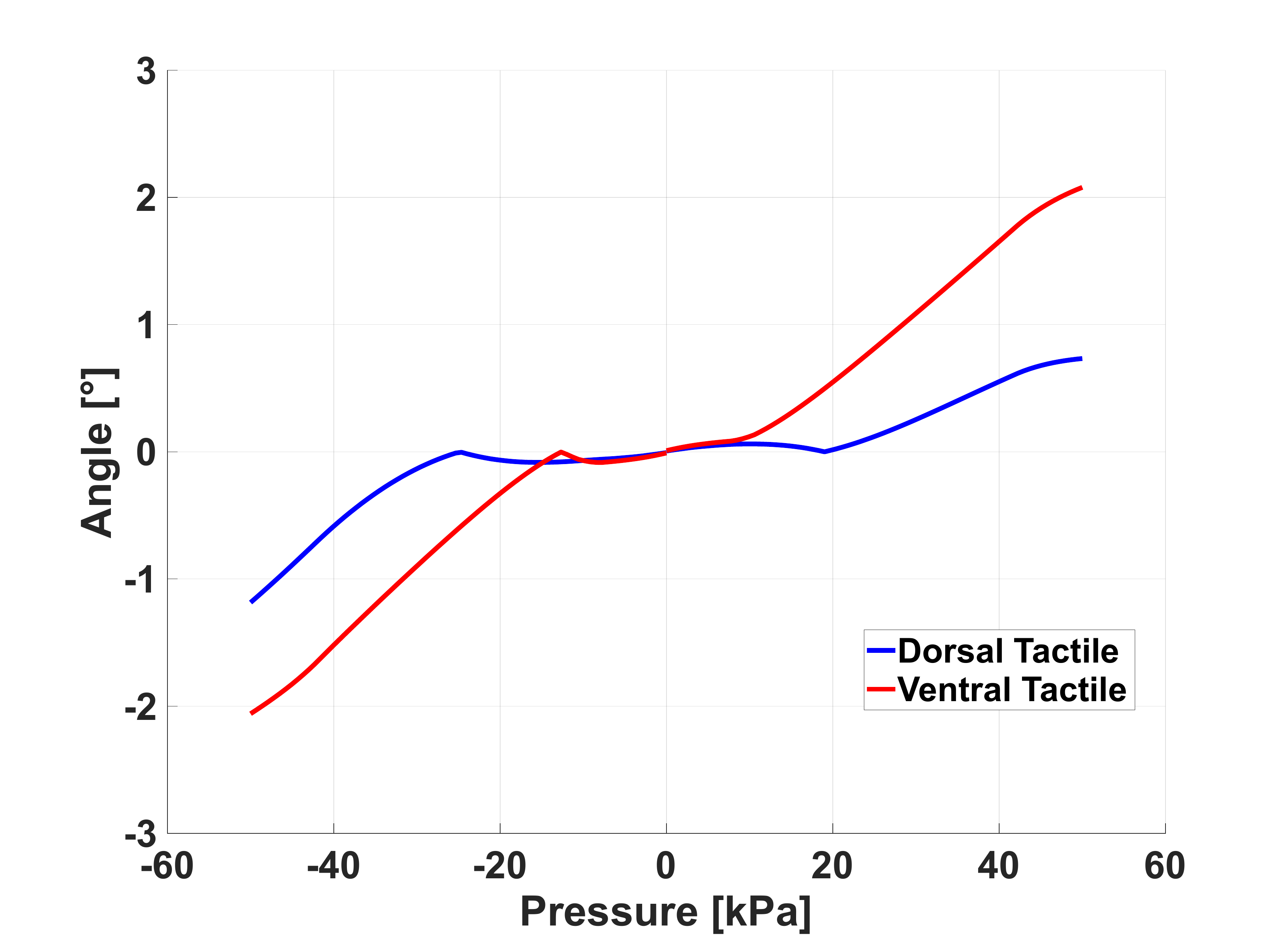}
    \caption{Angle--pressure relationships for Model~B obtained from Dorsal tactile sensor (blue) and Ventral tactile sensor (red).  
    The distinct responses illustrate the influence of sensor positioning on measured angular variations.}
    \label{fig:angle_sensors36}
\end{figure}

The angular response of integrated sensors was further analyzed for Model~B. In particular, the evolution of the angle with respect to the applied chamber pressure was computed for Dorsal and Ventral tactile sensors. The resulting curves are reported in Figure~\ref{fig:angle_sensors36}. 

\clearpage

\section*{Additional Application Information}

\begin{figure}[hbt!]
    \centering
    \includegraphics[width=0.5\linewidth]{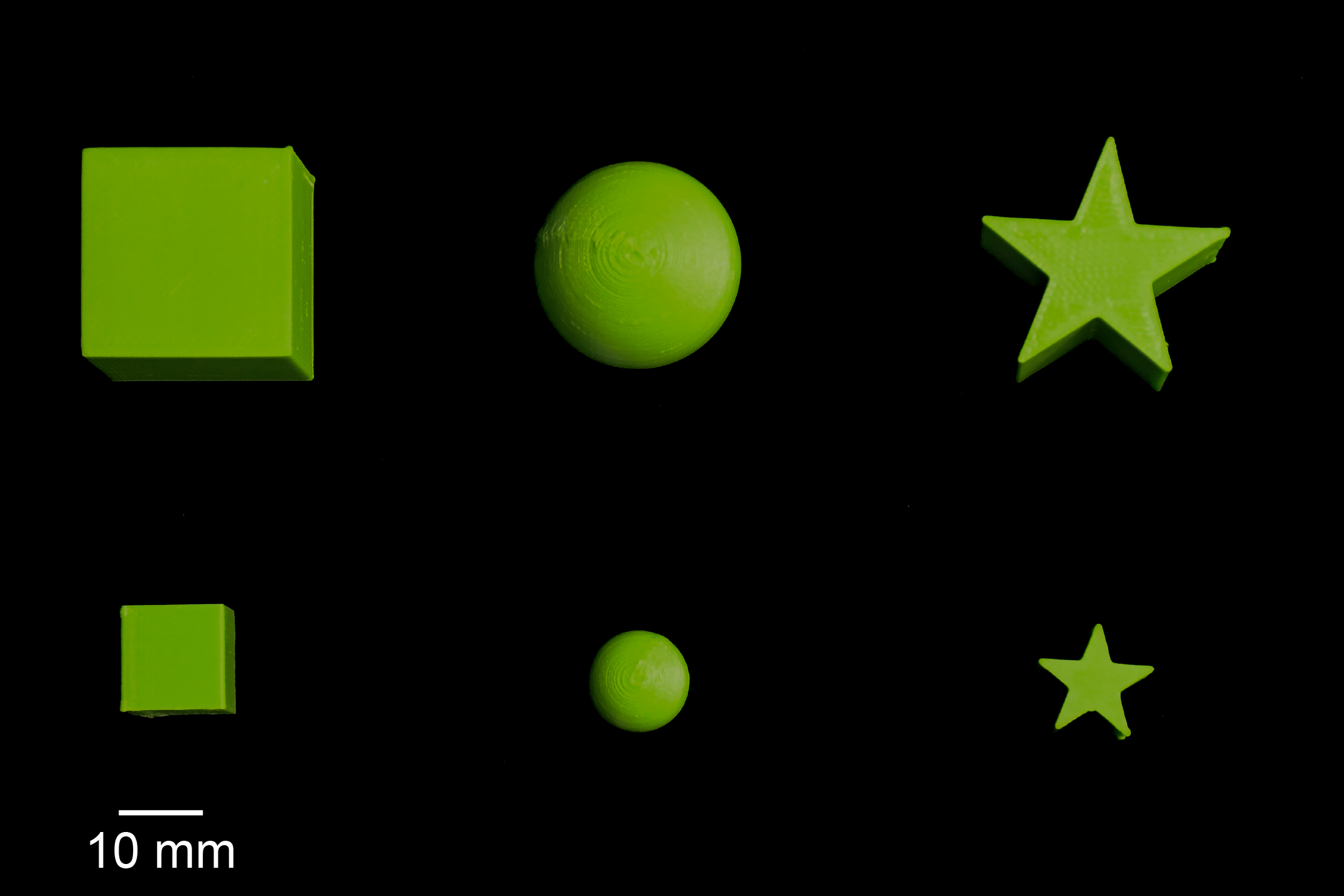}
    \caption{Cube, sphere, and star objects with sizes of (bottom) one unit cell (12.5 mm) and (top) two unit cells (25 mm).}
    \label{fig:objects}
\end{figure}

The objects that were grasped by MELEGROS in Video S3 are shown in Figure \ref{fig:objects}. All objects were printed with ABS with a desktop 3D printer (Ultimaker S7, Netherlands).

\bibliographystyle{MSP}
